\documentclass[11pt]{article}
\AtBeginDocument{%
  }

\usepackage[a4paper,margin=1in]{geometry}
\usepackage{amsmath,amssymb}
\usepackage{graphicx}
\usepackage{booktabs}
\usepackage{multirow}
\usepackage{hyperref}
\usepackage{subcaption}
\usepackage{makecell}
\usepackage{soul}
\usepackage{longtable}
\usepackage{lscape}
\usepackage[table]{xcolor}
\usepackage{xcolor}

\title{Longitudinal Multimodal Sensing of Physical Activity and Well-Being in Older Adults}
\author{
Flavio Di Martino\textsuperscript{1} \\ \and
Mattia G. Campana\textsuperscript{1}, \\ \and
Marcello Magno\textsuperscript{2}, \\ \and
Lorenza Pratali\textsuperscript{2}, \\ \and
Franca Delmastro\textsuperscript{1}
\\ \and
\textsuperscript{1}IIT-CNR, Pisa, Italy \quad
\textsuperscript{2}IFC-CNR, Pisa, Italy
}

\begin{document}
\maketitle




\begin{abstract}
Wearable and mobile sensing technologies enable continuous monitoring of human behavior and health in real-world settings.
However, predictive modeling in longitudinal multimodal data remains challenging, particularly when targeting complex or clinically derived outcomes.
In this work, we present a longitudinal multimodal study of 66 older adults conducted in real-world conditions and combining wearable sensing, behavioral monitoring, and clinical assessments. This setting provides a rare opportunity to study an underrepresented population in long-term, into-the-wild conditions.
Building on this dataset, we investigate how the alignment between sensed signals and target variables affects predictive performance across health-related tasks. We design a unified evaluation framework spanning tasks with increasing levels of observability, including \emph{Activity Levels} prediction, \emph{Sleep Duration} estimation, and \emph{Sleep Apnea Severity} classification.
Our results reveal a clear gradient of predictability: highly observable behavioral targets achieve robust performance (macro-F1 $\approx 65\%$), while more abstract outcomes remain challenging despite consistent improvements over baseline models. Moreover, through explainability analysis, we show that historical features consistently emerge as the most informative predictors, highlighting the central role of longitudinal information.
\end{abstract}




\section{Introduction}

The widespread adoption of wearable and mobile sensing technologies has opened new opportunities for monitoring human behavior and health in real-world environments~\cite{10.1371/journal.pmed.1001953, harari2016using}.
Continuous streams of physiological, behavioral, and contextual data enable the study of daily routines and long-term health trajectories at an unprecedented scale.
In particular, longitudinal multimodal sensing has emerged as a key paradigm for understanding how lifestyle factors, such as physical activity and sleep, influence well-being over time~\cite{CORNET2018120, Xu2022GLOBEM}.

However, despite significant progress, current research is still affected by several structural limitations.
First, most existing datasets are biased toward young or working-age populations, while older adults, who represent a primary target for preventive healthcare and long-term monitoring, remain largely underrepresented~\cite{Liu2021-fo}.
This limitation is particularly critical, as models trained on younger cohorts often fail to generalize to older populations, where physiological responses, behavioral patterns, and health conditions differ substantially.
Second, although many datasets provide high-resolution raw sensor signals, they often lack a consistent abstraction layer that enables scalable and interpretable longitudinal analysis~\cite{Vaizman2017ExtraSensory}.
In contrast, wearable-derived behavioral features, aggregated over meaningful time scales (e.g., hourly, daily, or weekly), offer a more suitable representation for modeling health and well-being, as they better align with the temporal dynamics of the underlying phenomena while enabling robust long-term analysis.
Finally, existing datasets rarely combine all desirable properties simultaneously, such as \emph{multimodality}, \emph{long-term longitudinal coverage}, \emph{in-the-wild collection}, and \emph{integration with clinically relevant outcomes}, thus limiting both modeling capabilities and scientific insights~\cite{Mattingly2019Tesserae, Mundnich2020TILES, Xu2022GLOBEM}.

In this work, we address these challenges through a comprehensive longitudinal study of older adults conducted entirely in real-world conditions.
We collected a multimodal dataset from 66 participants (mean age $72.6 \pm 4.4$ years) over approximately 23 months, spanning multiple experimental phases and combining continuous sensing, structured physical intervention, and repeated clinical assessments.
Participants were equipped with an integrated sensing ecosystem including a \emph{smartwatch}, an \emph{under-mattress sleep tracker}, a \emph{smart scale}, and a custom \emph{mobile application}, enabling the collection of heterogeneous data streams such as physical activity, cardiovascular signals, sleep dynamics, body composition, contextual information, and self-reported well-being measures.
The resulting dataset provides a detailed representation of daily life, collected entirely into-the-wild, capturing both short-term variability and long-term behavioral trends.

Due to the sensitive nature of the data and GDPR regulations, access to the dataset will be granted upon signing a Data Transfer Agreement (DTA) between the receiving institution and the authors’ institution, only for research purpose. The interested researchers could send a direct request to the authors including a brief introduction of the research project, and they will receive all the instructions to complete the necessary legal steps and access the data.
The dataset structure and characteristics are designed to support reproducible research and future extensions. A detailed description of the dataset, its metadata and the instructions for data access request are available at the following link https://aathe-ds.github.io/ (currently maintaining anonymity for review purpose). 

A distinctive aspect of our study is the integration of passive sensing with a structured \emph{Adapted Physical Activity} (\emph{APA}) program for older adults and periodic clinical evaluations.
The APA program represents both an intervention and an incentive for the involved subjects, providing important guidelines for physical activity that subjects can maintain in their daily life to improve their well-being. This design ensures a high user engagement, a psychological motivation in the use of the technology and allows the observation of natural behavioral patterns, in relation to clinically grounded outcomes, such as functional capacity and sleep-related conditions.
Moreover, the longitudinal nature of the dataset supports the extraction of historical features, such as lagged values and rolling statistics, that capture temporal dependencies and behavioral routines, providing a richer representation than instantaneous measurements alone.

Building on this dataset, we investigate the extent to which multimodal wearable-derived signals can support the prediction of health-related outcomes at different levels of abstraction.
Rather than focusing on a single task, we adopt a unified perspective based on the concept of \emph{observability}, i.e., the degree to which a target variable is directly reflected in the available sensing modalities.
To this end, we design a set of machine learning tasks that span a spectrum of increasing difficulty: (i) binary next-day physical activity prediction (active vs. non-active), representing a highly observable behavioral signal; (ii) binary sleep duration prediction, which reflects an intermediate and indirectly observable construct; and (iii) sleep apnea severity prediction, formulated as a three-class classification problem (mild, moderate, severe) based on clinically defined thresholds.

Our results reveal a clear and consistent gradient of predictability across these tasks.
For the most observable setting, activity prediction achieves robust performance (balanced accuracy and macro-F1 around $65\%$), despite the use of a strict Leave-One-Subject-Out (LOSO)  cross-subject evaluation protocol, which requires models to generalize across heterogeneous individuals.
For sleep duration, performance remains above baselines but exhibits increased variability, reflecting the more complex relationship between wearable signals and sleep-related outcomes.
The most challenging task is sleep apnea severity prediction, where models must discriminate between three clinically defined classes under weak signal-to-label alignment and inherent label ambiguity. In this setting, models achieve balanced accuracy above $50\%$ and macro-F1 around $50\%$, representing substantial improvements over both random ($+33\%$) and majority-class baselines.
Importantly, these results should be interpreted in light of the intrinsic difficulty of the task: the target variable is not directly observable from wearable signals, but instead emerges from complex physiological processes that are only partially captured by the sensing modalities.

Overall, our findings suggest that predictive performance in longitudinal multimodal sensing is fundamentally constrained by the degree of alignment between measured signals and target constructs, rather than by model complexity alone. This perspective provides a principled framework for interpreting both successful and challenging prediction tasks, highlighting the importance of considering observability when designing and evaluating machine learning approaches for health monitoring.

In summary, this work makes three main contributions:

\begin{enumerate}
    \item introduces a longitudinal multimodal dataset of older adults collected in-the-wild, combining wearable sensing, mobile data, and clinical assessments over an extended time span;

    \item provides a systematic characterization of data availability, quality, and temporal alignment in real-world conditions, demonstrating the feasibility of long-term monitoring in this population;

    \item proposes and validates an observability-driven framework for analyzing predictive performance across heterogeneous health-related tasks, offering insights into both the potential and the inherent limitations of wearable-based inference.
\end{enumerate}

The rest of the manuscript is organized as follows. Section~\ref{sec:related} reviews related work, with a particular focus on longitudinal and multimodal sensing datasets, highlighting their key characteristics and limitations.
Section~\ref{sec:study} describes the study design and presents the proposed dataset, detailing the population, sensing ecosystem, data collection protocol, and integration architecture.
Section~\ref{sec:dataset_overview} provides a comprehensive analysis of the dataset, including data availability, validity, and cross-modal alignment, followed by both population-level insights and machine learning experiments designed to assess predictive capabilities across tasks with different levels of observability (Section~\ref{sec:ml_exps}).
Finally, Section~\ref{sec:conclusions} concludes the paper, discussing the implications of our findings, outlining potential use cases of the dataset, and highlighting directions for future research.

\section{Background and Related works}
\label{sec:related}

Longitudinal multimodal sensing has emerged as a key paradigm for studying behavioral, physiological, and contextual dynamics in real-world environments. Recent datasets such as Tesserae \cite{Mattingly2019Tesserae}, TILES-2018 \cite{Mundnich2020TILES}, StudentLife~\cite{Wang2014StudentLife}, and GLOBEM~\cite{Xu2022GLOBEM} have demonstrated the feasibility of continuously collecting heterogeneous data streams, including wearable physiological signals, smartphone-derived behavioral features, and ecological momentary assessments (EMAs), over extended periods. These efforts highlight the potential of multimodal sensing for capturing real-world behavioral dynamics, while also exposing challenges related to temporal variability, data sparsity, and population heterogeneity.

In parallel, recent advances in wearable sensing have shifted from unimodal activity recognition toward multimodal learning frameworks that integrate heterogeneous sensor modalities. While early human activity recognition (HAR) systems were primarily based on inertial measurements, recent work increasingly leverages multimodal inputs, including physiological signals, contextual data, and cross-device sensing to improve robustness and representation quality~\cite{10.1145/3749502}.
For instance, recent studies explore multimodal wearable sensing and representation learning to better capture complex behavioral patterns and improve generalization across settings and populations~\cite{Xu2022GLOBEM, 10.1145/3328932}.
These approaches highlight the importance of combining complementary sensing modalities to address noise, ambiguity, and variability in real-world data.
However, despite these advances, most multimodal HAR approaches remain focused on short-term activity recognition tasks and do not explicitly address longitudinal behavioral dynamics. In particular, the datasets used in these works are often limited in duration and do not capture long-term temporal dependencies, which are critical for modeling routines, lifestyle patterns, and health trajectories.

Recent multimodal datasets such as ExtraSensory \cite{Vaizman2017ExtraSensory} and Lifetrace \linebreak\cite{Lifetrace2021} represent a significant step toward real-world deployment, combining wearable sensing, contextual information, and in-the-wild data collection. In particular, Lifetrace provides longitudinal multimodal data collected over extended periods, integrating wearable and contextual signals to study daily behavior in natural settings. However, these datasets still predominantly focus on young and healthy populations. While they improve ecological validity and multimodal richness, they present key limitations: data collection is often constrained in duration compared to real-world longitudinal monitoring, and clinically relevant ground truth remains limited.

As a result, longitudinal multimodal datasets often provide limited representation of older adults, even though this population has the greatest need for continuous, long-term monitoring and health evaluation.
These limitations have direct implications for predictive modeling on longitudinal multimodal data. Existing works tend to focus on highly observable outcomes, such as physical activity levels or coarse behavioral states, where the mapping between sensor signals and labels is relatively direct. In contrast, predicting more abstract or clinically derived constructs, such as sleep quality or health conditions, is inherently more challenging due to weaker alignment between observed signals and underlying phenomena, as well as increased label uncertainty. This gap is further exacerbated by the limited availability of reliable ground truth for such outcomes.
Furthermore, the longitudinal nature of these datasets enables the use of historical features, such as lagged observations, rolling statistics, and exponentially weighted summaries, which capture temporal dependencies and behavioral routines. While these approaches can improve predictive performance, their effectiveness depends strongly on the observability of the target and the temporal variability of the data. In relatively stable or weakly informative settings, historical summaries may capture general trends but provide limited discriminative power for fine-grained prediction.
Overall, existing datasets tend to satisfy only a subset of desirable properties—longitudinality, multimodality, in-the-wild deployment, wearable sensing, and population diversity—but rarely all simultaneously. In particular, the combination of long-term multimodal sensing, real-world deployment, and a focus on older adult populations remains underexplored.

In this work, we address these challenges by leveraging a longitudinal multimodal dataset collected in older adults under real-world conditions, and by systematically analyzing predictive tasks across different levels of observability. In particular, we investigate how the alignment between sensor signals and target variables, as well as the use of historical features, affects predictive performance across behavioral and clinically derived outcomes.

\section{Pilot study and Dataset}
\label{sec:study}

\begin{figure}[t]
    \centering
    \includegraphics[width=0.99\linewidth]{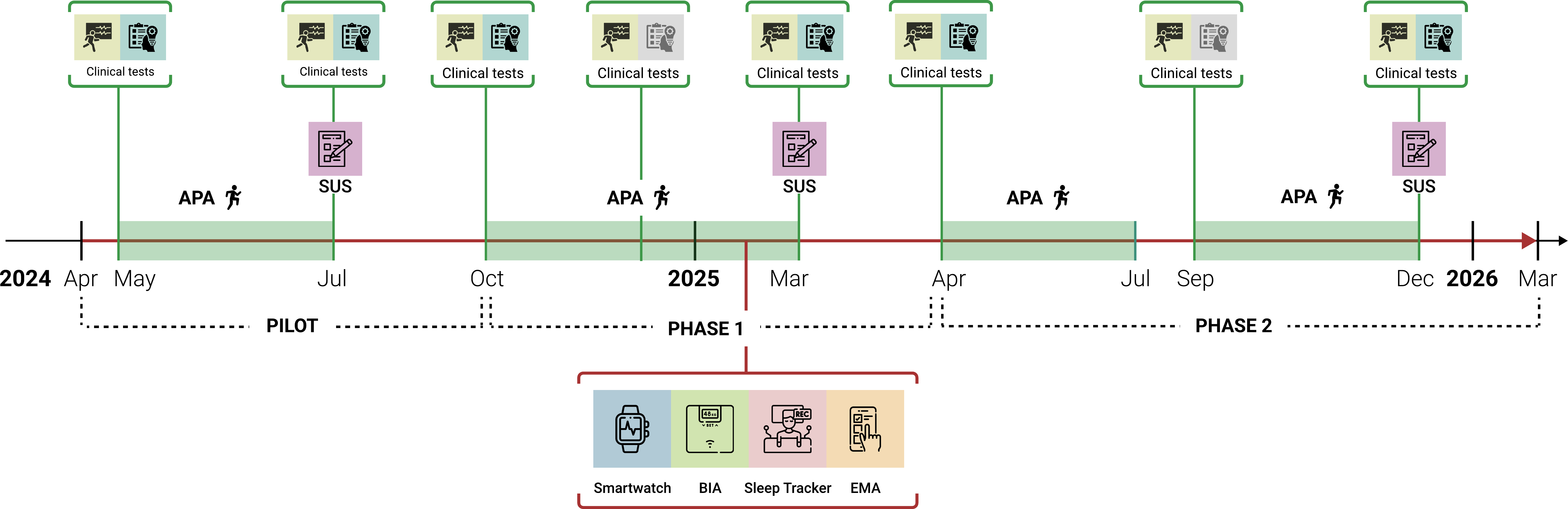}
    \caption{Experimental protocol.}
    \label{fig:protocol}
\end{figure}

The study is the result of the multidisciplinary collaboration among computer scientists, cardiologists, sport scientists and personal trainers. 
The main idea originated by a clinical need to promote physical activity among older adults and to better understand its impact on health and well-being over time. Through a strict collaboration we designed a monitoring framework capable of supporting both intervention and assessment, by leveraging longitudinal and multimodal data collected in real-world settings. This need is aligned with international recommendations, such as the World Health Organization (WHO) guidelines on physical activity for older adults \cite{WHO2020}, which recommends 150–300 minutes of moderate physical activity per week. This roughly corresponds to an additional 2,000–4,000 steps per day that, when combined with habitual daily activity, translates to approximately 6,000–8,000 total steps per day in older adults. Although step count is a convenient and widely adopted indicator of physical activity, it captures only a limited aspect of individual health, motivating the need to integrate additional behavioral, physiological, and sleep-related dimensions.
Within this context, our study aims to combine continuous monitoring of multiple domains with predictive modeling to capture how behavioral, physiological, and sleep-related signals evolve over time and reflect changes in health status.

To support this objective, we implemented a structured Adapted Physical Activity (APA) program, delivered by certified personal trainers in an instrumented gym facility coordinated by the authors’ institution. The program was offered over a period of six months to three consecutive groups, with sessions held twice per week, thus serving both as the intervention phase and as an incentive for sustained participation in the study. Each user received a sensing ecosystem that was configured through the personal smartphone and directly at home by researchers, as detailed in Section \ref{sub:ecosystem}.
During the training sessions, the personal trainers provided personalized exercise programs to each subject, tailored to the individual characteristics. The program would also serve as guidance aimed at promoting the continuation of physical activity outside the APA setting.

As illustrated in Figure~\ref{fig:protocol}, the study ran from the end of April $2024$ to the end of March $2026$, lasting approximately $23$ months divided into three phases: \texttt{\textbf{Pilot}}, \texttt{\textbf{Phase 1}}, and \texttt{\textbf{Phase 2}}. The Pilot phase was designed as a testing phase, both for the technological solutions and for the user engagement. In each phase a different group of users have been involved. They attended two training APA sessions per week (1 hour each), which included tailored exercise programs designed to improve functional capacity and promote independent living through safe, supervised physical activities. APA, as defined by the International Federation of Adapted Physical Activity \cite{IFAPA}, focuses on mobility, balance, strength, and cardiovascular health, while adapting activities to individual capabilities and limitations.
We maintained a consistent session schedule across all trials, with interruptions due to summer closures of the gym facility or users' holiday periods (e.g., from July to September $2024$ during the Pilot, and from July to August $2025$ within Phase $2$), as well as transitions between consecutive trials.
Nevertheless, active users were allowed to continue using the sensing ecosystem at home and in their free time for remote monitoring, although full adherence during these periods has been generally reduced.

The Pilot spanned from the end of April to the end of September $2024$, with APA intervention conducted for approximately two months during May and June. It provided an initial feasibility and deployment validation stage, supporting the verification of device provisioning, cloud synchronization, service adherence, and backend integration procedures.
Phase $1$ started in October $2024$ and continued until the end of March $2025$, representing the first long-term monitoring period under stable deployment conditions, covering six months of combined remote sensing and intervention. Finally, Phase $2$ extended from April $2025$ to March $2026$, with an intermediate summer break in the APA program during July and August.
Across all trials, participants also attended three clinical test sessions for \texttt{baseline} (T$0$), \texttt{mid-term} (T$1$), and \texttt{follow-up} (T$2$) assessment\footnote{Please note that the T$1$ assessment was not performed in the Pilot trial due to the limited duration of this phase.}.
At each session, participants underwent a list of clinical evaluations in order to provide a complete anamnesis, including functional capacity levels, cognitive status, malnutrition, sleep disorders, and emotional states (see Table \ref{tab:screening_tests}. 

\begin{table}[h!]
\small
\centering
\caption{List of clinical tests.}
\label{tab:screening_tests}
\resizebox{\textwidth}{!}{%
\begin{tabular}{llcc}
\toprule
\textbf{Test} & \textbf{Acronym} & \textbf{Target Domain} & \textbf{Output} \\
\midrule
Mini Mental State Examination~\cite{MMSE}& MMSE& Cognitive status & 0-30\\
\midrule
Mini Nutritional Assessment~\cite{MNA}& MNA& Nutritional status & 0.30 \\
\midrule
Pittsburgh Sleep Quality Index~\cite{PSQI}& PSQI & Sleep Quality & 0-21\\
\midrule
Depression, Anxiety, and Stress Scales~\cite{DASS-21}& DASS-21 & Mood & 0-21 \\
\midrule
International Physical Activity Questionnaire~\cite{IPAQ}& IPAQ & Self-reported Physical activity Level & MET-minutes/week\\
\midrule
6 Minutes Walking Test~\cite{6MWT}& 6MWT & Functional Capabilities & Distance covered (meters) \\
\midrule
Timed Up and Go~\cite{TUG}& TUG & Mobility/Fall risk & Execution time (s)\\
\bottomrule
\end{tabular}}
\end{table}
In addition, physiological evaluations have been conducted in terms of body composition, ecocardiography, blood pressure and oxygenation.
Ambulatory blood pressure measurement was made  wearing a portable device connected to an armband that automatically measures blood pressure at regular intervals (three readings taken on the left arm, one minute apart) \cite{Zhang2025ABPM}.

The repeated clinical evaluations provided an anchor for long-term remote monitoring, enabling an objective and clinically-grounded interpretation of the subjects' physio-behavioral trajectories over time.
Eventually, at the end of each trial, participants also completed the \texttt{System Usability Scale} (SUS) to evaluate the
acceptability and usability of the proposed technological solutions. In addition, all the participants accepted to be recontacted after 6 months and 1 year from the end of the study for a follow-up evaluation. 

As a result, the overall study design combines structured physical intervention, continuous sensing, and clinical evaluations. Its multi-phase approach also supports both cross-sectional comparisons between different subject cohorts and within-subject longitudinal analyses spanning both in- and out-of-protocol periods. The study received Institutional Review Board (IRB) approval from the Ethics Committee of the authors’ institution (full details will be provided upon acceptance to preserve anonymity), and all participants provided written informed consent.

\subsection{Population}

\begin{table}[t]
\small
\centering
\caption{Population characteristics.}
\label{tab:participants}
\resizebox{\textwidth}{!}{%
\begin{tabular}{llrrrr}
\toprule
\textbf{Category} & \textbf{Variable} & \textbf{Overall} & \textbf{Pilot} & \textbf{Phase 1} & \textbf{Phase 2} \\
\midrule
\multirow{4}{*}{Demographics} 
& Participants (N) & 66 & 20 & 29 & 30 \\
& Gender: Female & 41 (62.1\%) & 8 (40.0\%) & 17 (58.6\%) & 22 (73.3\%) \\
& Gender: Male & 25 (37.9\%) & 12 (60.0\%) & 12 (41.4\%) & 8 (26.7\%) \\
& Age (years) & 72.6 $\pm$ 4.4 & 73.2 $\pm$ 3.9 & 72.5 $\pm$ 4.5 & 72.3 $\pm$ 4.3 \\
\midrule
\multirow{9}{*}{Physiological} 
& Body Mass Index - BMI (Kg/m$^2$)& 25.8 $\pm$ 4.0 & 26.2 $\pm$ 3.8 & 26.7 $\pm$ 4.3 & 25.2 $\pm$ 3.8 \\
& Fat Mass (\%) & 31.3 $\pm$ 7.6 & 31.0 $\pm$ 8.1 & 31.3 $\pm$ 8.7 & 31.4 $\pm$ 6.6 \\
& Fat Mass (Kg) & 22.3 $\pm$ 7.3 & 23.8 $\pm$ 8.3 & 22.8 $\pm$ 7.9 & 21.8 $\pm$ 6.8 \\
& Fat-Free Mass - FFM (Kg) & 47.8 $\pm$ 8.7 & 52.8 $\pm$ 11.7 & 48.7 $\pm$ 9.8 & 47.0 $\pm$ 7.7 \\
& Total Body Water - TBW (Kg) & 35.0 $\pm$ 6.4 & 38.7 $\pm$ 8.6 & 35.7 $\pm$ 7.2 & 34.4 $\pm$ 5.7 \\
& Heart Rate - HR (BPM) & 70.2 $\pm$ 11.4 & 65.7 $\pm$ 8.6 & 67.8 $\pm$ 10.7 & 72.5 $\pm$ 11.8 \\
& Systolic Blood Pressure - SBP (mmHg) & 148.7 $\pm$ 21.8 & 149.3 $\pm$ 21.4 & 147.8 $\pm$ 23.4 & 149.5 $\pm$ 20.5 \\
& Diastolic Blood Pressure - DBP (mmHg) & 79.9 $\pm$ 10.9 & 80.8 $\pm$ 6.4 & 79.0 $\pm$ 9.7 & 80.6 $\pm$ 11.9 \\
& Peripheral Blood Oxygenation - SpO$_2$ (\%) & 96.7 $\pm$ 1.3 & 96.9 $\pm$ 1.1 & 96.4 $\pm$ 1.7 & 97.1 $\pm$ 0.8 \\
\bottomrule
\end{tabular}}
\end{table}

Table~\ref{tab:participants} summarizes the demographic, as well as the baseline anthropometric and physiological characteristics of the study population. Overall, we enrolled a total of $N=66$ older adults. Specifically, $20$ subjects participated in the Pilot, $29$ in Phase $1$, and $30$ in Phase $2$. An overlap of $13$ subjects occurred between Pilot and Phase $1$, while Phase $2$ consisted exclusively of new enrollments. This is related to the limited duration of the pilot phase from which we decide to maintain the subjects who demonstrated a high engagement and adherence both to the APA program and the remote monitoring. 
From a demographic perspective, the overall sample was predominantly female, comprising $41$ women ($62.1\%$) and $25$ men ($37.9\%$). However, the gender distribution varied between phases: the Pilot and Phase $1$ cohorts were relatively balanced (approximately $60–40\%$), while Phase $2$ was characterized by a clear predominance of women ($73.3\%$). This pattern is consistent with previous evidence indicating greater participation of women in structured physical activity interventions among older adults.

In terms of age distribution, participants had a mean age of $72.6 \pm 4.4$ years, reflecting a homogeneous older population, as further supported by the comparable mean age across phases and their narrow standard deviations.
A similar pattern of comparability was observed for the remaining characteristics. Statistical testing using Welch’s t-test did not reveal significant differences between Phase $1$ and Phase $2$, nor between Phase $2$ and the Pilot phase. 

Participants were generally in the upper range of normal weight or slightly overweight according to standard BMI categories~\cite{Winter2014}. Mean systolic and diastolic blood pressures were $148.7 \pm 21.8$ mmHg and $79.9 \pm 10.9$ mmHg, respectively, consistent with values commonly observed in older populations, where hypertension is prevalent \linebreak\cite{Garg2018-zn}. Some cases could also be associated with white coat hypertension for the presence of a healthcare professional.
Finally, peripheral capillary oxygen saturation (SpO$_2$) remained high and stable across phases ($96.7 \pm 1.3\%$), suggesting healthy respiratory function at baseline.

\subsection{Sensing ecosystem}
\label{sub:ecosystem}
To enable continuous and unobtrusive monitoring of participants in real-world conditions, we adopted a sensing ecosystem based on commercial off-the-shelf devices, thus ensuring ecological validity, measurement reliability, and scalable data management. 
The participants were equipped with a \textbf{\emph{Garmin Venu Sq 2}} smartwatch to continuously monitor physiological and activity-related data.
It integrates optical photoplethysmography (PPG) sensors to estimate HR, SpO\textsubscript{2}, \emph{stress levels} derived from heart rate variability (HRV), as well as \emph{breathing rate} (BR). In addition, it includes an accelerometer to capture activity-related measures such as \emph{step count} and \emph{activity intensity}.

Sleep-related metrics were collected using a \textbf{\emph{Withings Sleep Analyzer}}, an unobtrusive under-mattress device designed for long-term monitoring in home environments.
It features a pneumatic sensor that measures HR, BR, and body movements via ballistocardiography (BCG), as well as a sound sensor that detects audio signals associated with breathing interruptions. The device provides a detailed sleep summary, including (but not limited to) \emph{sleep stages} (i.e., awake, light, deep, REM), number of awakenings and out-of-bed events, physiological parameters, as well as \emph{snoring} measures. Importantly, it also estimates the \emph{Apnea–Hypopnea Index} (AHI), a clinically validated, medical-grade indicator representing the average number of apneas (complete breathing pauses) and hypopneas (shallow breathing events) events per hour of sleep, which is a particularly valuable indicator for assessing sleep-related breathing disturbances.
Body weight and composition were periodically assessed using a \textbf{\emph{Garmin Index S2}}, with a recommended measurement frequency of at least $3$ times per week.
In addition to \emph{body weight}, the scale performs bioelectrical impedance analysis (BIA) to estimate body composition parameters, including \emph{fat mass}, \emph{muscle mass}, \emph{bone mass}, and \emph{total body water}. It is worth noting that the clinical measurements of body composition relies on a certified medical device, reporting thus a more accurate measurements, but the commercial smart scale allows for repeated measurements at home.
Overall, all these devices offer validated sensing capabilities widely adopted in consumer health ecosystems, ensuring stable and reliable measurements in real-world conditions. Moreover, they leverage mature cloud infrastructures for data synchronization, storage, and secure access, which greatly facilitates large-scale longitudinal data collection and remote monitoring. By integrating with the vendors’ official cloud architectures, we ensured continuous data availability while minimizing participant burden and technical maintenance. All data were automatically synchronized with participants’ smartphones via the Garmin Connect and Withings HealthMate companion apps and uploaded to the vendors’ cloud infrastructures for secure retrieval and analysis.
The sensing data collection procedure is detailed in the next subsection. Notably, once installed, the sleep tracker and smart scale operate passively by sending their data via the user’s home WiFi network, reducing participant burden and promoting adherence. On the other hand, smartwatch management requires minimal active user interaction, mainly for periodic recharging. 

In addition to the devices described above, we deployed a custom-developed native mobile application (for both Android and iOS) for active and passive data collection. The former consisted of daily \emph{Ecological Momentary Assessments} (EMAs) covering \emph{physical activity}, \emph{social interactions}, \emph{mood}, and \emph{dietary habits}. 
These self-reported data are primarily intended to provide fine-grained, day-level ground truth for downstream analyses, but they can also serve as additional contextual features. For each participant, we scheduled a daily reminder notification after dinner, according to individual preferences, to prompt EMA completion. The app also provided a summary dashboard displaying recent data from each device, enhancing transparency and participant engagement. On the other hand, passive sensing includes periodic \emph{human activity recognition} (HAR) via on-device physical and virtual sensors, and \emph{location} tracking, in order to complement physiological and behavioral data. Both features rely on Google/Apple APIs, with daily readings transmitted at the end of each day through a background task scheduled within the mobile app. 
The sensing ecosystem has been deployed before each pilot phase. Researchers installed and configured the required mobile applications on participants' personal smartphones, and provided the wearable devices. The installation and configuration of the remaining monitoring devices required a dedicated deployment at participants’ homes, as these required access to the domestic WiFi network.


\begin{figure}[t]
    \centering
    \includegraphics[width=0.75\textwidth]{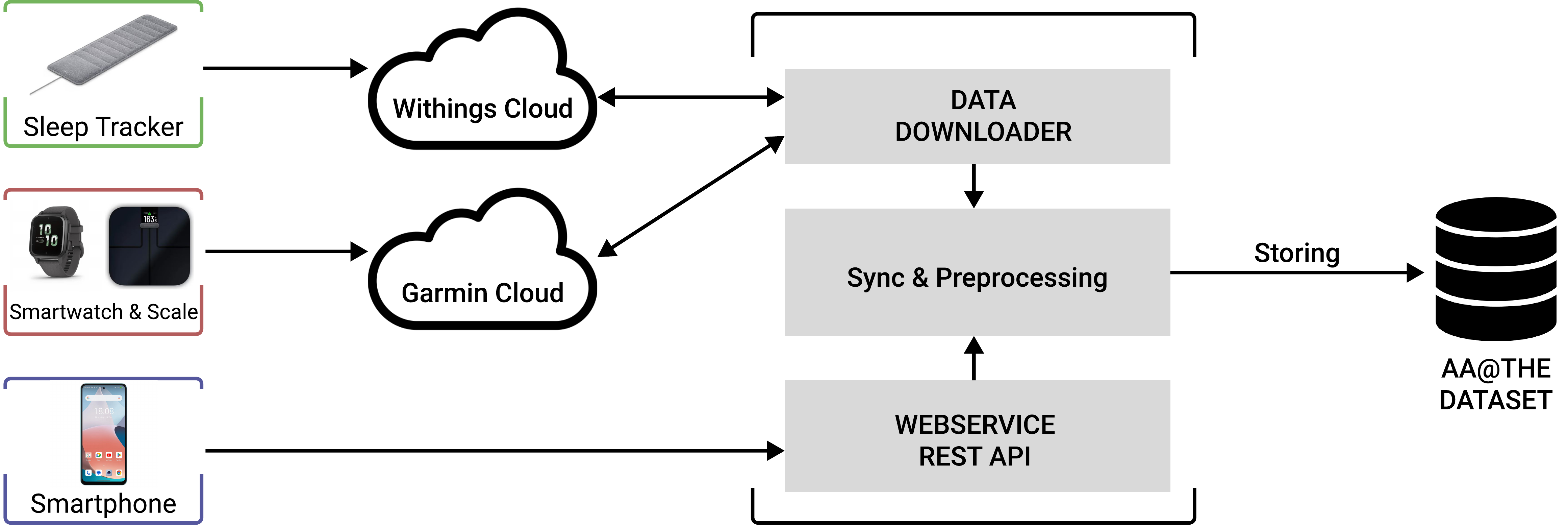}
    \caption{Data collection architecture}
    \label{fig:data_collection_arch}
\end{figure}

Figure~\ref{fig:data_collection_arch} illustrates the overall data collection architecture adopted in our study. 
The infrastructure was specifically designed to (i) maintain a clear separation between device-specific vendor ecosystems 
and the research backend, (ii) guarantee secure data transmission, and (iii) enable daily automated integration of heterogeneous 
data streams into a unified research dataset. 
For each participant, we created a unique, fully anonymized account to serve as the sole identifier across the entire infrastructure, without any reference to the personal data of the associated user. 
These IDs were used to set up dedicated accounts on the Garmin\footnote{\url{https://connect.garmin.com}} and Withings\footnote{\url{https://app.withings.com}} cloud platforms 
and to initialize the corresponding devices. The same identifier was embedded in our custom mobile application to ensure consistent 
cross-platform data linkage.
This approach embeds participant privacy into the system design itself.

From a data communication perspective, the Garmin smartwatch and smart scale 
automatically sync with the Garmin cloud, while the sleep tracker uploads data to the Withings cloud upon detecting wake-up. 
These daily synchronizations leverage vendor-managed infrastructures for secure transmission, storage, and availability, 
providing industrial-grade security, reliable service, and robust authentication while minimizing operational and maintenance 
demands on the research system.
To bridge the gap between vendor cloud platforms and our research repository, we implemented an automated server-side component, 
referred to as the \emph{Data Downloader} in Figure \ref{fig:data_collection_arch}. A daily scheduled task programmatically 
authenticates to the Garmin and Withings cloud services using the anonymized user accounts to retrieve newly available records.
In parallel with vendor-mediated data flows, the custom smartphone application used a dedicated, fully controlled transmission pathway
over HTTPS, ensuring encrypted end-to-end transfer of sensitive information. 


\section{Dataset validation}
\label{sec:dataset_overview}
\begin{figure}[t]
    \centering
    \includegraphics[width=0.99\textwidth]{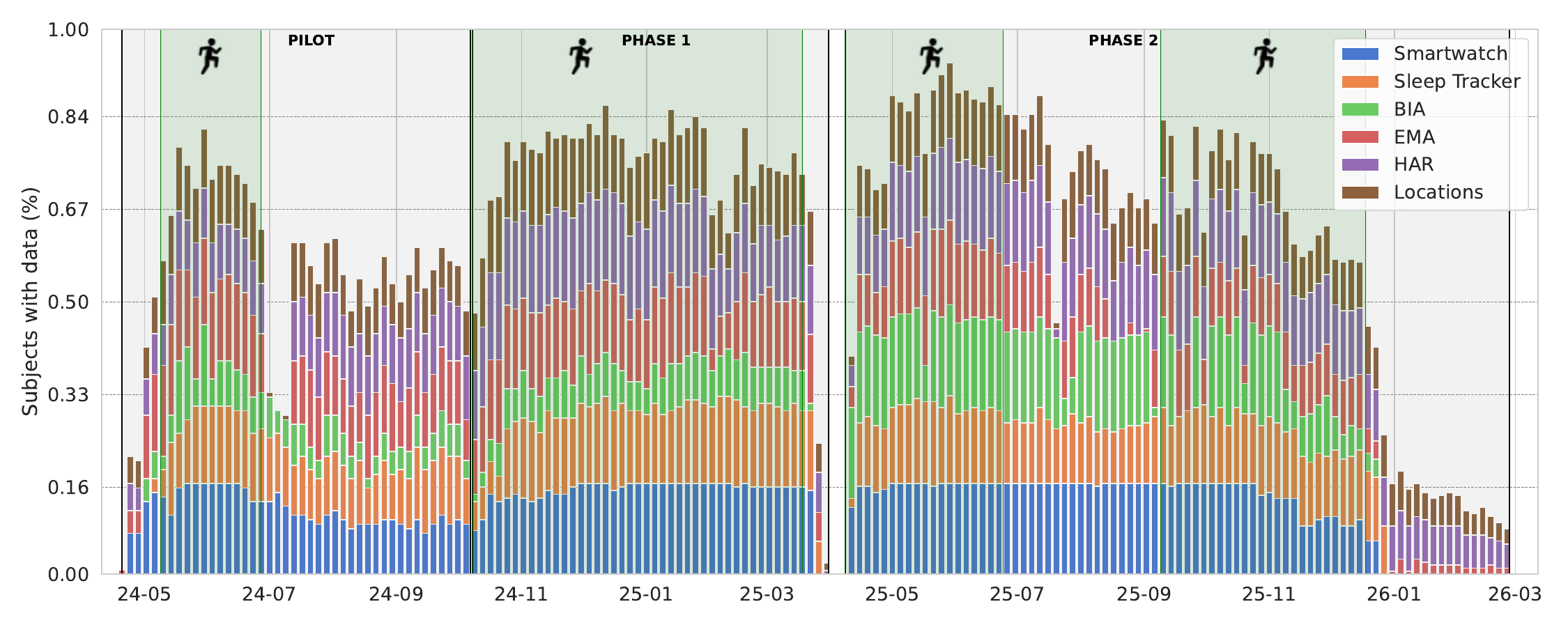}
    \caption{Dataset overview throughout the study duration.}
    \label{fig:dataset_overview}
\end{figure}

Figure \ref{fig:dataset_overview} provides an overview of the daily data collected throughout the study. The colored bars indicate the percentage of available data from each of the six sensing modalities, namely smartwatch (SW), sleep tracker (ST), scale (BIA), EMA, and smartphone-based HAR and location data (LOC). The three consecutive monitoring trials lasted \texttt{\textbf{170}}, \texttt{\textbf{173}}, and \texttt{\textbf{267}} days, respectively. This preliminary assessment of daily availability assumes that each user contributes at least with one sample per modality and is expressed relative to the number of active participants in each trial. 
In the ideal case, when all participants provide data for all modalities, the bars sum to $1.0$. It is important to note that only a small number of dropouts has been experienced in the entire study, due to personal reasons or sudden health issues. Specifically, only $4$ subjects from Phase $1$ have been excluded resulting in a final cohort of $25$ participants for this trial ($6\%$ drop out on the entire pilot study).

As we can observe, data availability is highest during APA periods across all trials, with daily percentages consistently above $75\%$ and all sensing modalities represented in most cases. This was expected since APA represented also the incentive to increase the users' engagement. 
However, data from several sources remain available during non-APA periods as well, with the exception of the gap between Phase $1$ and Phase $2$, which is due to the deployment transition between two different user groups. In addition, a long tail of data is observed after December 2025, resulting from delays in retrieving the sensing devices, which led to continued data collection from a subset of volunteer users. This final phase was excluded from the evaluation process.

In terms of study phases, we can note a decrease in the central part of Phase $2$, which is related to the summer break between the $2$ internal APA periods, in which users spent some time on holiday and the hot climate contributed to slightly reduce the engagement in physical activity. 
It is worth specifying that participants were instructed to use their devices continuously throughout the APA period, while they were free to use the devices for the rest of the time. Results highlight that most of the users maintained their engagement also outside the APA periods, providing important data for the long-term analysis.
To this aim, we compared the data availability in the full trial with only APA periods that result in \texttt{\textbf{50}} days for the Pilot, \texttt{\textbf{162}} days for Phase$1$, and \texttt{\textbf{179}} days for Phase $2$. Specifically, we analysed device-specific data by defining the following custom criteria as validity thresholds. We defined an assessment window from 8:00 A.M. to 8:00 P.M. for the smartwatch use, and  we considered valid those days that have at least $75\%$ of the expected HR measurements within the window, based on the HR sampling rate, which corresponds to a minimum wearing time of $8$ hours.

For scale data, a day was considered valid if both weight and body composition parameters have been registered. In fact, in some cases we found only partial measurements due to incorrect use of the scale (e.g., wearing socks or shoes, or standing on the scale for a limited time). When multiple scale measurements were recorded in a day, we used their average values.
For sleep tracker data and EMA, only complete data transmission are allowed by design. Therefore, if we have data for a day, they are intrinsically valid.

\begin{figure}[t]
    \centering
    \begin{subfigure}[t]{0.32\textwidth}
        \centering
        \includegraphics[width=\linewidth]{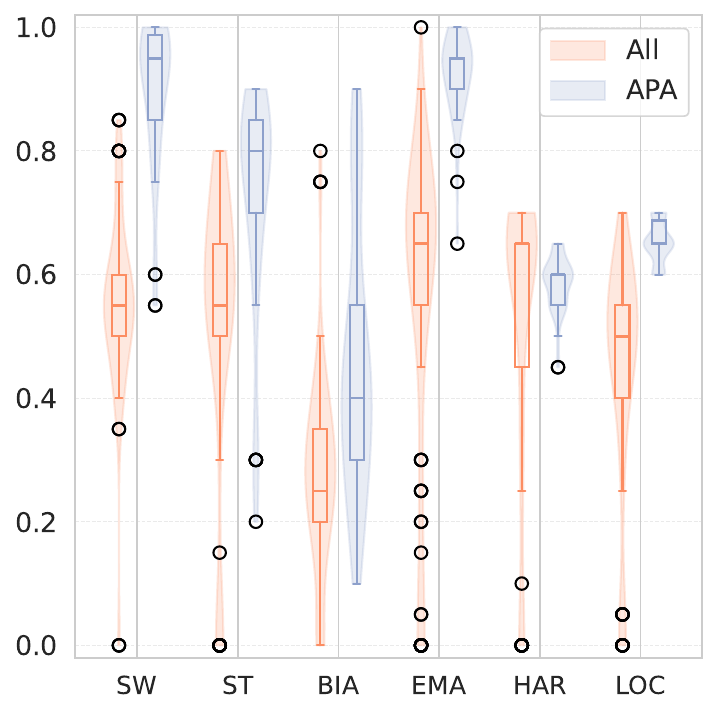}
        \caption{Pilot}
    \end{subfigure}%
    ~ 
    \begin{subfigure}[t]{0.32\textwidth}
        \centering
        \includegraphics[width=\linewidth]{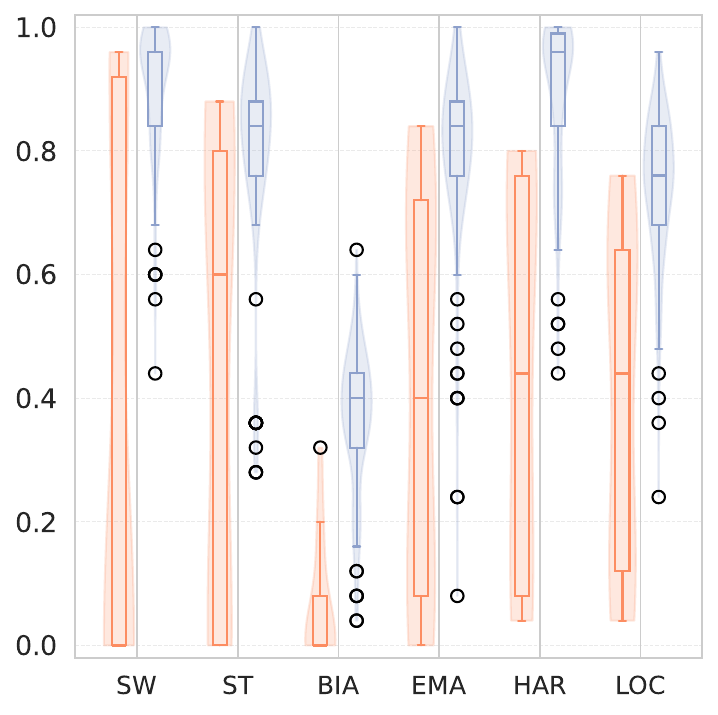}
        \caption{Phase 1}
    \end{subfigure}
    ~ 
    \begin{subfigure}[t]{0.32\textwidth}
        \centering
        \includegraphics[width=\linewidth]{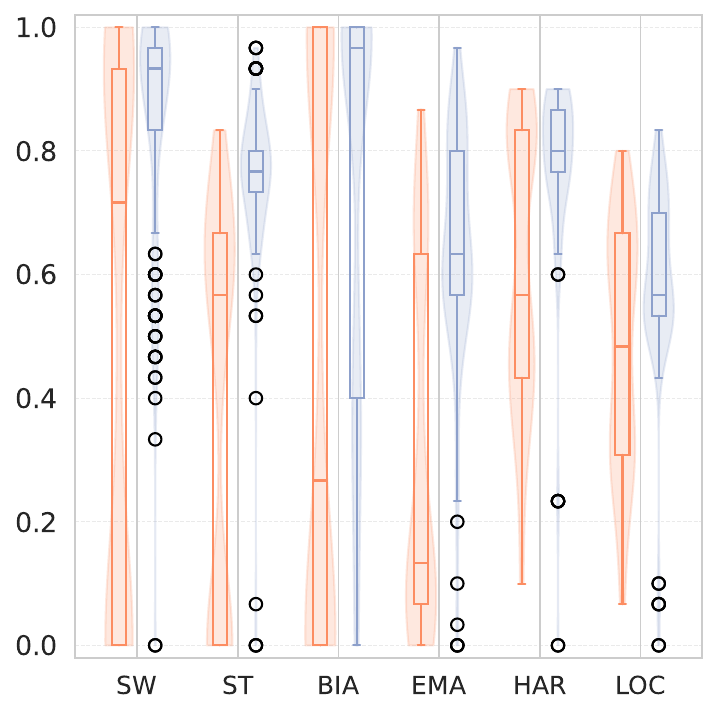}
        \caption{Phase 2}
        \label{fig:validity_phase2}
    \end{subfigure}
    \caption{Comparison of data validity distributions between full (All) and APA periods across all trials.}
    \label{fig:full_vs_apa_data_validity}
\end{figure}

Figure \ref{fig:full_vs_apa_data_validity} reports the distribution of daily validity rates across sensing modalities and trials. Each value represents the proportion of valid daily samples aggregated across users within a given modality, visualized via violin and box plots for both APA and full study observation windows.
In general, as expected, the APA periods consistently represent the most reliable observation window across sensing modalities, yielding higher median validity and reduced dispersion, thereby reflecting optimal conditions in which participants are actively engaged and device usage is most controlled. This is particularly evident for smartwatch and smartphone-based sensing, where APA boundaries capture stable high-validity regimes across trials. However, while a degradation in data quality naturally occur when extending the analysis outside the incentive phase, most data remain usable. Some particular cases are reduced validity are related to EMAs and BIA, mainly due to the opportunistically use of the related devices. Considering this situation we can support the use of the full-period data for longitudinal analysis, yet with increased missingness due to less controlled nature of real-world data collection.

\begin{figure*}[t]
    \centering
    \begin{subfigure}[t]{0.33\textwidth}
        \centering
        \includegraphics[width=\linewidth]{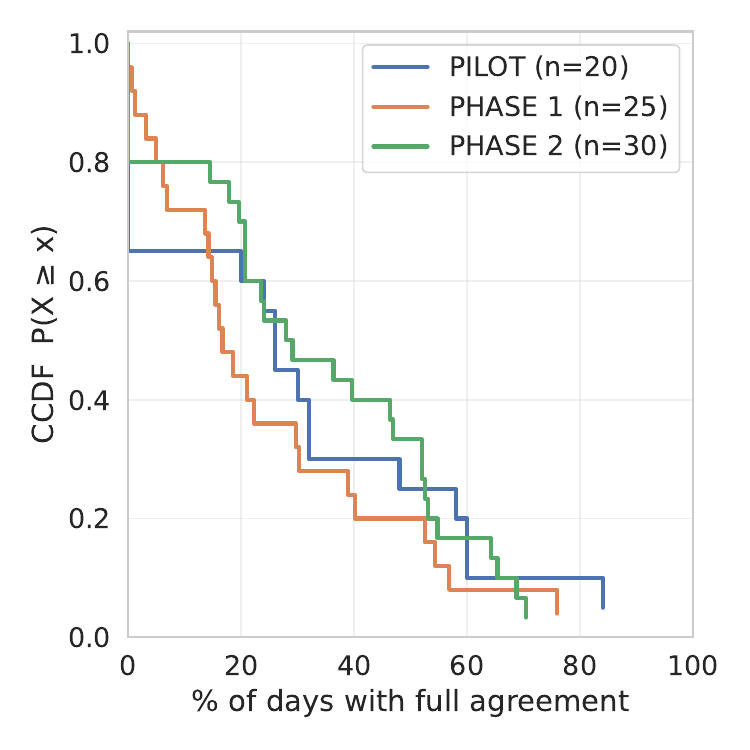}
        \caption{All daily modalities}
    \label{fig:agreement_all}
    \end{subfigure}
    ~
    \begin{subfigure}[t]{0.33\textwidth}
       \centering
     \includegraphics[width=\linewidth]{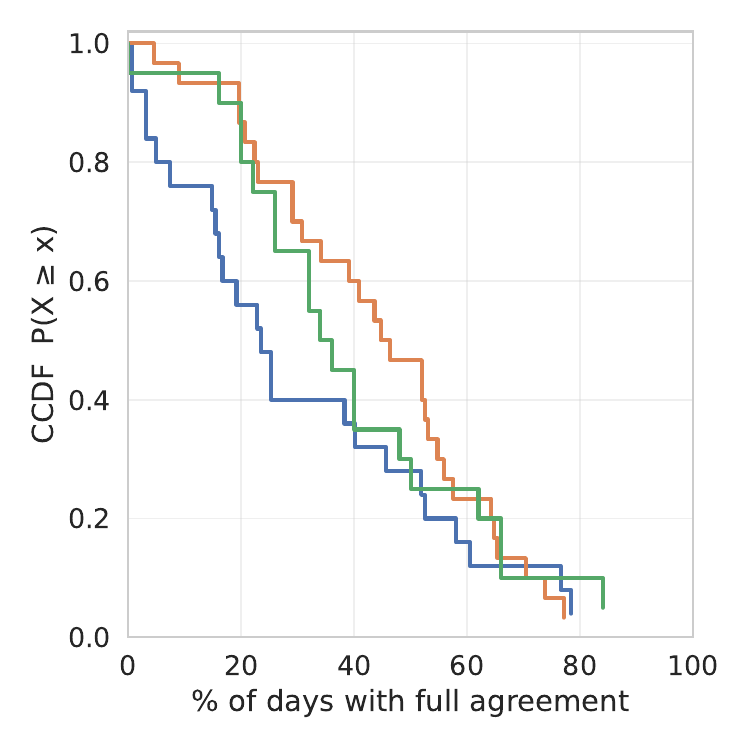}
    \caption{All W/o smartphone-embedded sensing}
    \label{fig:agreement_smartphone}
    \end{subfigure}
    ~
    \begin{subfigure}[t]{0.33\textwidth}
        \centering
        \includegraphics[width=\linewidth]{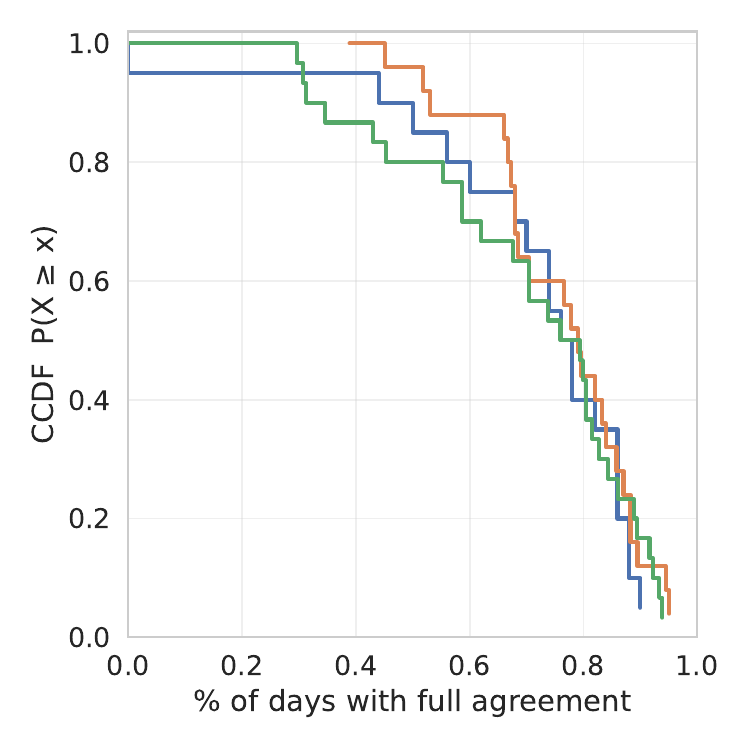}
        \caption{Smartwatch and Sleep Tracker}
        \label{fig:smartwatch_sleep_agreement}
    \end{subfigure}
     ~
  
    \caption{CCDF visualizations of cross-modal data availability.}
    \label{fig:agreement_cross_device}
\end{figure*}

\subsection{Cross-Modal Data availability and temporal alignment}
\label{sec:agreeement_analysis}

To assess cross-modal data availability and temporal alignment, we defined a daily completeness indicator for each user and trial, capturing whether all expected sensing modalities provided valid data within the same day. We then quantified the proportion of days with complete multimodal coverage across all sources. We originally discarded BIA value due to the limited number of measurements.

To characterize the distribution of this measure across participants, we analyzed its complementary cumulative distribution function (CCDF), which quantifies the fraction of users achieving at least a given percentage of days with complete cross-modal data availability within each trial. This analysis provides a global view of multimodal data coverage, highlighting the extent to which data from different sensing sources are temporally aligned and jointly available for downstream analyses.

Figure \ref{fig:agreement_cross_device} illustrates the overall cross-modal data availability and alignment across different combinations of sensing devices and modalities.
Specifically, Figure \ref{fig:agreement_all} shows cross-modal data availability when considering all daily sensing modalities, namely smartwatch, sleep tracker, EMAs, and smartphone-based location and activity data. The distribution reveals substantial heterogeneity across users. At relatively low thresholds, most users satisfy the criterion: approximately $90\%$ of users achieve at least $20\%$ of days with complete cross-modal data availability. As the threshold increases, the proportion of users progressively decreases. About $70\%$ of users achieve at least $40\%$ of such days, while this proportion drops to approximately $45$–$50\%$ at a threshold of $60\%$. Overall, these results indicate moderate data completeness when considering the joint availability and temporal alignment of all daily modalities.

The situation changes if we remove smartphone-embedded sensing, and even more considering two main devices: smartwatches and sleep tracker.
In the first case, the corresponding CCDFs are shifted upward compared to those obtained when all daily modalities are considered. The situation further improves in Figure \ref{fig:smartwatch_sleep_agreement}, indicating high completeness levels between the smartwatch and the sleep tracker. More than $80\%$ of the users provides complete data on at least $60\%$ of days, and roughly $60-70\%$ of users go over $80\%$ of days. The distributions are broadly consistent across the three trials, with only minor variations. Overall, these results support the combined use of activity and sleep data for reliable full-day monitoring of the participants. 

\begin{figure*}[t]
\centering
    \begin{subfigure}[t]{0.32\textwidth}
        \centering
        \includegraphics[width=\linewidth]{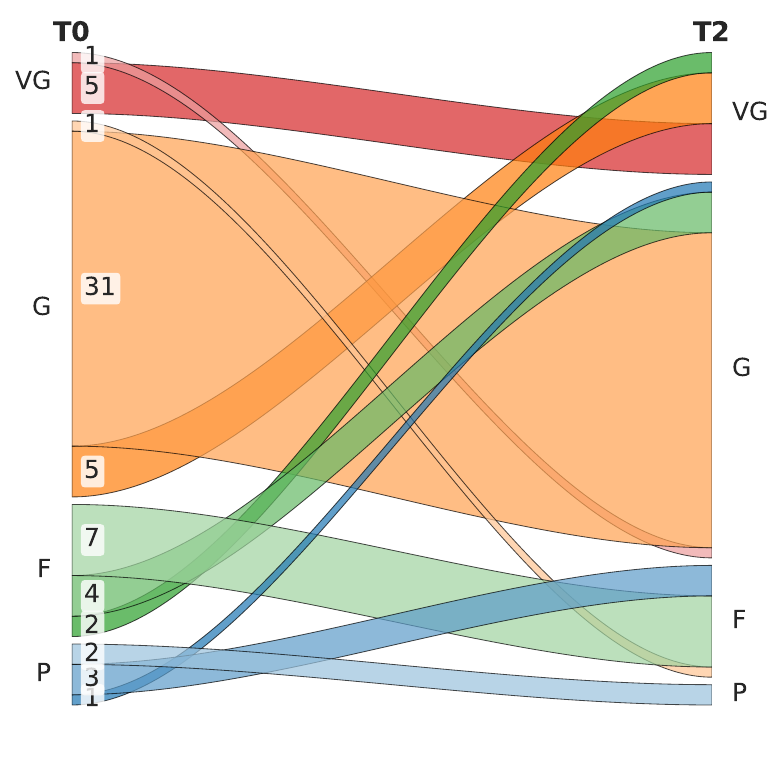}
        \caption{FC changes between T0 and T2}
        \label{fig:sankey}
    \end{subfigure}
    ~
    \begin{subfigure}[t]{0.32\textwidth}
        \centering
        \includegraphics[width=\linewidth]{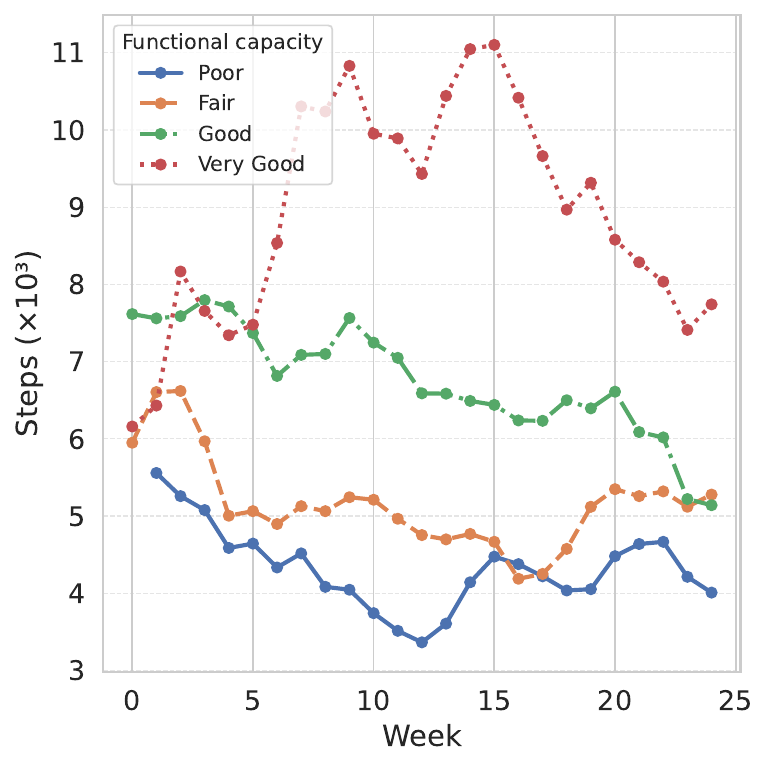}
        \caption{Step count}
        \label{fig:steps}
    \end{subfigure}
    ~ 
    \begin{subfigure}[t]{0.32\textwidth}
        \centering
        \includegraphics[width=\linewidth]{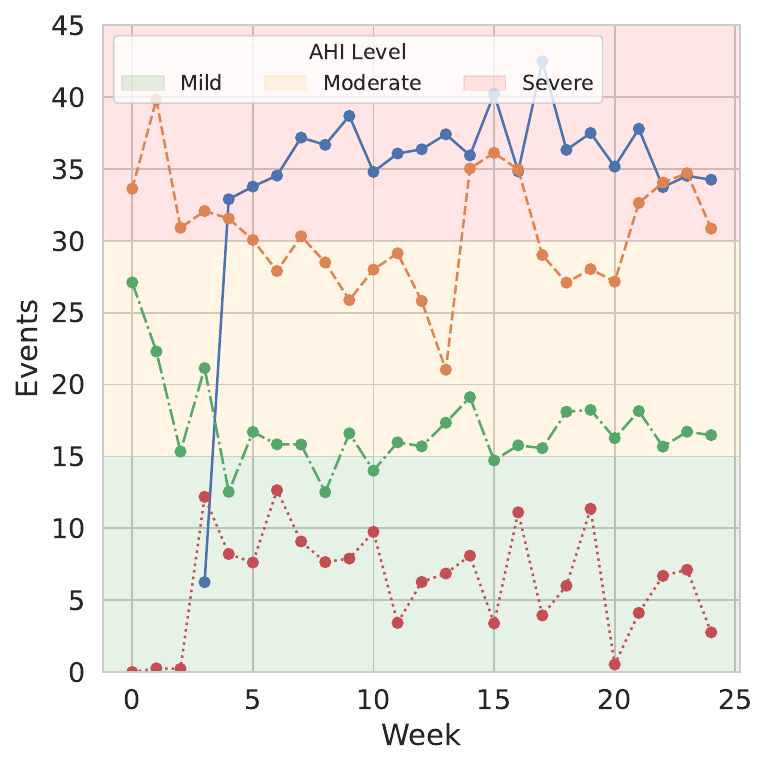}
        \caption{AHI}
        \label{fig:ahi}
    \end{subfigure}
    \caption{Changes in users' functional capabilities from baseline to follow-up assessment (a). Average weekly trends during Phase $1$ for step count (b) and AHI (c) features, stratified by users’ baseline FC categories (derived from 6MWT).}
    \label{fig:trends}
    
\end{figure*}

\section{Population Stratification and Multimodal Signal Analysis}

Starting from the clinical assessments collected at T$0$ and T$2$, we analyzed the impact of the APA program on older adults’ well-being, both as a structured intervention and as a driver for adopting healthier lifestyles.
To this end, we considered two key dimensions of the clinical evaluation: functional capacity (FC), measured through the Six-Minute Walk Test (6MWT) \cite{6MWT}, and Body Mass Index (BMI). The 6MWT requires each participant to walk for six minutes along a flat, straight path, and the total distance covered (in meters) is used to classify subjects into four FC levels: \textit{Poor} (distance $<300$ m), \textit{Fair} ($300 \leq$ distance $\leq 400$ m), \textit{Good} ($400 <$ distance $\leq 500$ m), and \textit{Very Good} (distance $>500$ m).
First, we stratified all participants based on FC, using the evaluations at T$0$ and T$2$, and analyzed the evolution of each subject across categories, as illustrated in Figure~\ref{fig:sankey}.
The Sankey diagram reveals a predominantly stable trend among participants with \textit{Good} functional capacity: $45$ out of $62$ individuals maintained their initial FC level. In the context of aging populations, such stability represents a clinically meaningful outcome, as preserving functional capacity over time is itself indicative of positive health status. In addition, $15$ participants showed a clear improvement, including $5$ subjects who achieved a two-category improvement, while only $2$ individuals experienced a slight decline, mainly due to concurrent health issues. Overall, these trends suggest a beneficial effect of the APA program, characterized by both the maintenance of functional capacity in the majority of participants and measurable improvements in a substantial subset of the population, with minimal deterioration observed.

To further characterize these groups, we analyzed the temporal evolution of behavioral signals in terms of daily steps, and derived sleep-related outcomes, with particular attention to the Apnea–Hypopnea Index\footnote{AHI is a clinically defined index computed as the average number of apnea and hypopnea events per hour of sleep, generally based on multiple physiological signals collected during polysomnography.} (AHI), across FC and BMI strata. Daily step count provides a highly observable and reliable proxy of functional mobility and engagement in physical activity, which is directly measurable from wearable sensors and widely used in both clinical and real-world monitoring settings. In contrast, AHI represents a clinically meaningful but less directly observable outcome, reflecting sleep-related breathing disturbances that are associated with cardiovascular and overall health risks. By jointly analyzing these two variables, we aim to cover a spectrum of observability, from directly measurable behavioral signals to clinically derived indicators, enabling a more comprehensive characterization of health-related patterns in the population.

Temporal trend analyses are reported for Phase 1 only, as this period corresponds to a consistent intervention phase during which participants were regularly engaged in the APA program under stable conditions. In contrast, Phase 2 is characterized by a non-stationary structure, including an initial APA period, an intermediate summer break with no structured activity or incentives, and a subsequent reintroduction of the program.
This sequence introduces substantial variability in behavioral patterns, driven by both the temporary interruption of the intervention and seasonal effects, resulting in a not negligible decline in activity levels during the break period. Consequently, including Phase 2 in the trend analysis would confound the interpretation of temporal dynamics, as observed variations would reflect not only individual behavior but also externally induced changes in engagement conditions.
For this reason, we restricted the trend analysis to Phase 1, in order to ensure a more consistent and interpretable evaluation of the relationship between multimodal signals and functional capacity under stable intervention conditions (see Figures \ref{fig:ahi} and \ref{fig:steps}).

When stratifying by FC, we observe a clear and monotonic separation across all groups. Higher FC levels are consistently associated with higher step counts, with well-separated trajectories across the observation period. A similar pattern emerges for AHI, where participants in higher FC groups exhibit systematically lower values, while lower FC groups show progressively higher AHI levels. This results in a highly structured and stable separation across groups, with limited overlap and consistent ordering over time.
However, these findings should be interpreted with caution, particularly for AHI. As a derived clinical index, AHI is influenced by multiple factors beyond functional capacity, including age, BMI, and sex, and is therefore not directly determined by behavioral or functional measures alone. The strong separation observed across FC groups may thus reflect the fact that FC captures a broader latent health construct, which is correlated with these determinants, rather than indicating a direct causal relationship between functional capacity and sleep apnea severity.

These observations are further supported by statistical analysis of activity-related data. 
One-way ANOVA indicated a significant overall effect of FC group on daily step counts, with post-hoc pairwise comparisons showing significant differences across all group pairs (all adjusted $p < 0.01$). 
In particular, comparisons involving the \textit{Very Good} FC group showed large effect sizes, suggesting a marked separation from the other categories, although this result can be affected by the relatively small sample size of the group.
However, when accounting for inter-subject variability through linear mixed-effects models (LMEM), the strength of these differences is reduced. Using the lowest group (\textit{Poor FC}) as reference, only the contrast with the highest group remains statistically significant, while the intermediate groups show positive but non-significant differences. 
This suggests that, although activity levels broadly follow FC stratification, the most robust behavioral contrasts are primarily observed between the extreme groups, while adjacent categories exhibit more gradual transitions.
Overall, these results indicate that FC effectively captures differences in activity levels, while also highlighting that the separation between neighboring groups is limited, reflecting a continuous rather than sharply discretized functional spectrum.

This result is consistent with the concept of observability, as step count represents a highly observable behavioral signal closely aligned with functional capacity. While this leads to strong overall group differences, only the most distinct FC levels remain robustly separable when accounting for inter-subject variability.

\begin{figure*}[t]
\centering
     \begin{subfigure}[t]{0.33\textwidth}
        \centering
        \includegraphics[width=\linewidth]{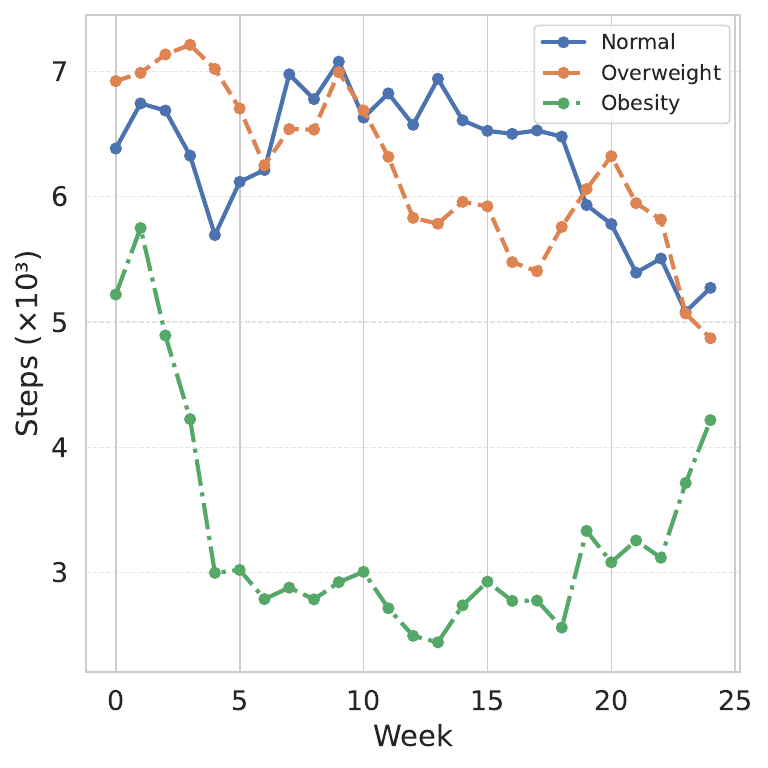}
        \caption{Step count}
        \label{fig:bmi-steps}
    \end{subfigure}
    ~ 
    \begin{subfigure}[t]{0.33\textwidth}
        \centering
        \includegraphics[width=\linewidth]{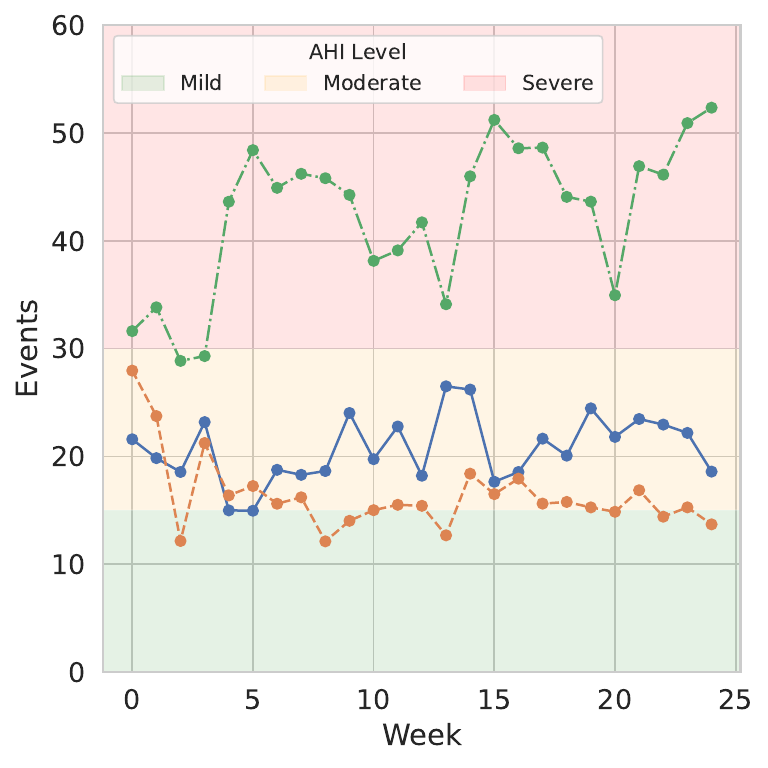}
        \caption{AHI}
        \label{fig:bmi-ahi}
    \end{subfigure}
    \caption{ Average weekly trends during Phase $1$ stratified by users’ baseline BMI.}
    \label{fig:bmi-trends}
    
\end{figure*}

In contrast, BMI-based stratification leads to more heterogeneous and less regular patterns (Figures~\ref{fig:bmi-steps} and \ref{fig:bmi-ahi}). 
While the obesity group consistently exhibits lower activity levels and higher AHI values—consistent with established clinical evidence—the separation between normal and overweight groups is less pronounced, particularly for activity. 
These observations are supported by statistical analysis. Post-hoc pairwise comparisons using Dunn’s test revealed significant differences across BMI groups for both daily step count and AHI, confirming a progressive reduction in activity with increasing BMI. However, linear mixed-effects models did not identify significant group-specific temporal slopes, indicating that while groups differ in their overall levels, their trajectories over time remain relatively stable.
A similar pattern is observed for AHI, where obese individuals consistently fall within the severe sleep apnea range, whereas normal-weight and overweight groups are predominantly distributed between mild and moderate levels. Also in this case, LMEM analysis did not reveal significant temporal interactions, suggesting that AHI behaves as a relatively stable, subject-specific characteristic over the observation period. While this is consistent with the known physiological nature of sleep apnea, the observed stability may also reflect the specific composition of the study cohort, and should therefore be interpreted with caution when generalizing to broader populations.
Overall, these results indicate that the apparent strength of relationships in multimodal sensing data depends critically on the choice of stratification variable. FC-based grouping provides stronger separability and clearer trends, but may partially reflect underlying confounding factors, while BMI-based grouping offers a more conservative and clinically grounded perspective, better capturing the inherent variability of physiological conditions.

Building on these observations, we next shift from stratified analysis to individual-level predictive modeling, where performance directly reflects the observability of each target, allowing us to evaluate how well multimodal signals can support outcome inference without relying on predefined group structures.

\subsection{Predictive modelling evaluation}
\label{sec:ml_exps}

To complement the stratified analysis, we formulate a set of predictive tasks aimed at quantifying the extent to which multimodal wearable signals can support the inference of health-related outcomes at the individual level.
Rather than treating each task independently, we adopt a unified perspective based on the notion of \emph{observability}, where target variables differ in how directly they are reflected in wearable sensor data.
We hypothesize that predictive performance is not solely driven by model complexity, but is primarily determined by the degree of alignment between the measured signals and the underlying latent construct.
To this end, we design a set of experiments spanning targets with progressively lower levels of observability, ranging from directly measurable behavioral signals (i.e., daytime activity levels), to moderately derived outcomes (i.e., sleep-related variables), up to clinically defined indicators (i.e., sleep severity proxies).


As a first step, we consider a binary \emph{activity classification} task, namely \textbf{Activity Levels}, where the goal is to forecast, one day ahead, whether a subject exceeds a predefined daily activity threshold.
Given healthy older adults as the reference population, we set a threshold of $6$K daily steps to distinguish between \texttt{active} and \texttt{non-active} subjects.
This task represents the most direct mapping between sensor signals and target labels, as physical activity is explicitly captured by wearable devices, particularly through smartwatch-derived measurements.
As such, it serves as a high-observability reference task, providing an upper bound on predictive performance when the target variable is strongly aligned with the available sensing modalities.

\begin{figure}[htbp]
    \centering
    \begin{subfigure}{0.32\textwidth}
        \centering
        \includegraphics[width=\linewidth]{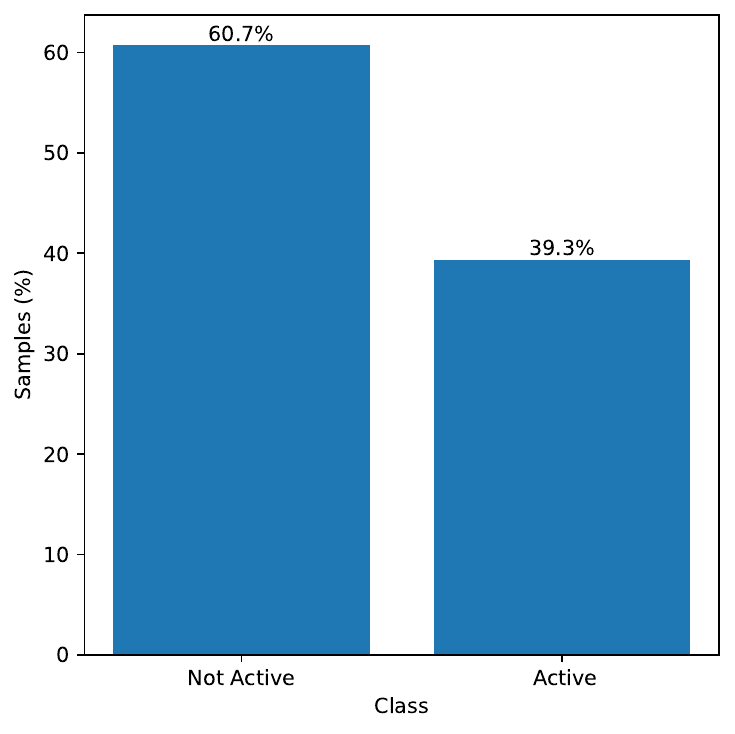}
        \caption{Activity Levels}
        \label{fig:activity_class_dist}
    \end{subfigure}
    \hfill
    \begin{subfigure}{0.32\textwidth}
        \centering
        \includegraphics[width=\linewidth]{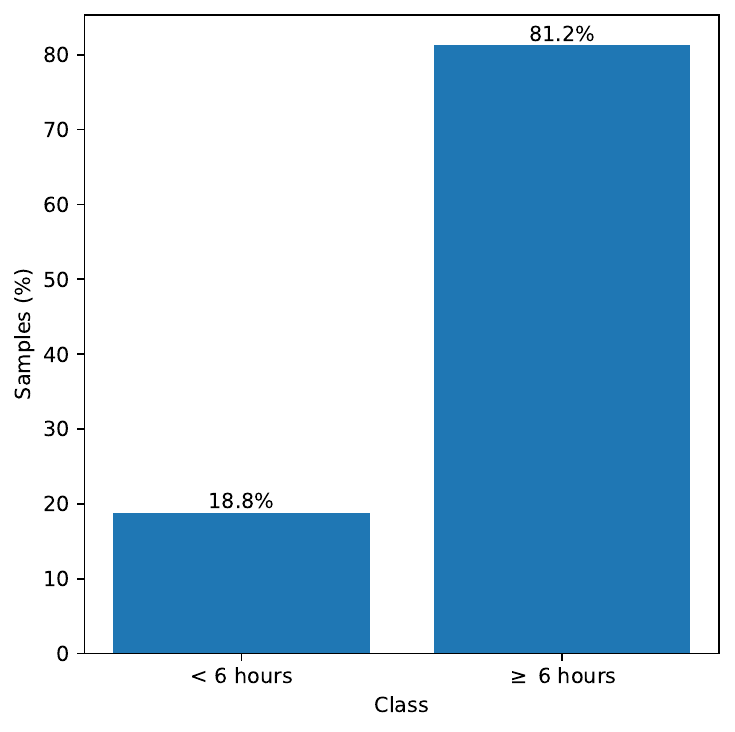}
        \caption{Sleep duration}
        \label{fig:sleep_dur_class_dist}
    \end{subfigure}
    \hfill
    \begin{subfigure}{0.32\textwidth}
        \centering
        \includegraphics[width=\linewidth]{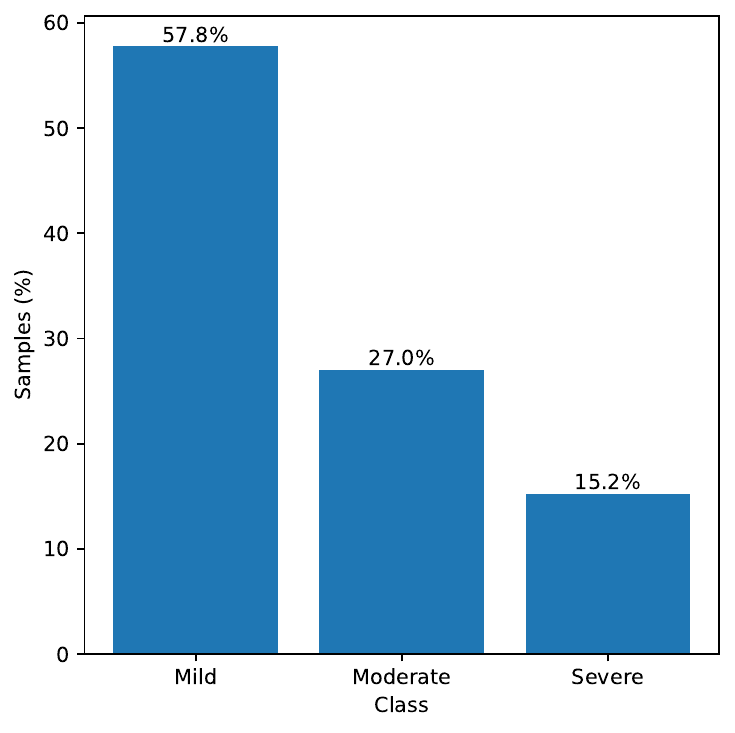}
        \caption{Sleep AHI}
        \label{fig:sleep_ahi_class_dist}
    \end{subfigure}
    \caption{Global class distributions across predictive tasks.}
    \label{fig:class_distributions}
\end{figure}

We then consider \textbf{Sleep duration} as a proxy for sleep quality, defining a binary prediction task based on whether the subject achieves a sufficient amount of sleep.
Unlike activity, sleep duration is not directly measured but inferred from multiple sensor-derived signals, making it inherently less observable.
While a conventional threshold of $8$ hours is typically used, we adopt a threshold of $6$ hours to better reflect the characteristics of the older population, where shorter sleep durations are more common \cite{sleep_duration}. This choice had a negligible impact on the class distribution, as using the higher threshold would result in only a $0.4\%$ difference with respect to the class proportions shown in Figure~\ref{fig:sleep_dur_class_dist}.


Finally, we consider the task \textbf{Sleep AHI} in which we evaluate the prediction of \emph{AHI severity}. According to clinical guidelines, it is categorized into three classes: \texttt{mild} (AHI $<15$), \texttt{moderate} ($15 \leq \mathrm{AHI} \leq 30$), and \texttt{severe} (AHI $>30$). This task introduces two key challenges. First, there is a weak coupling between wearable-derived signals and the underlying clinical condition, as sleep apnea is only indirectly reflected in the available measurements. Second, the discretization of a continuous physiological variable into fixed thresholds introduces label ambiguity, particularly near class boundaries. As a result, this task operates in a low-observability regime, where prediction is fundamentally constrained by limited signal-to-label alignment.

Across all tasks, we employ a consistent modeling and evaluation pipeline in order to isolate the effect of task observability from methodological variability. This allows us to interpret performance differences as a function of signal-to-label alignment, rather than dataset-or model-specific artifacts. Overall, our experimental design is intended to characterize a gradient of predictability across different constructs, ranging from strongly observable behavioral signals to weakly identifiable clinical abstractions, in order to better understand both successful and failure cases.

\begin{table}[t]
\centering
\caption{Summary of feature types used in the machine learning experiments.}
\begin{tabular}{lll}
\hline
\textbf{Data Type} & \textbf{Feature} & \textbf{Dimension} \\
\toprule
Steps & Daytime stats & 13 $\times$ 1 \\
HR  & Daytime stats & 9 $\times$ 1\\
RR &Daytime stats & 8 $\times$ 1\\
Stress & Daytime stats& 8 $\times$ 1\\
SpO$_2$ & Daytime stats& 8 $\times$ 1\\
Sleep & Sleep summary + derived feature & 41 $\times$ 1\\
\midrule
Lagged & Values over previous 3 days & 3 $\times$ feature\\
Rolling & Window stats (3 and 7 days) &  24 $\times$ window $\times$ feature\\
EWM & Window stats (3 and 7 days) & 3 $\times$ window $\times$ feature\\
Static & Age, Gender, BMI & 3 $\times$ 1\\
Contextual & Calendar-derived & 14 $\times$ 1\\
\bottomrule
\end{tabular}
\label{tab:ML_features}
\end{table}

\subsubsection{Feature engineering and extraction}

The \textbf{Activity Levels} task focuses on predicting the next-day physical activity level using heterogeneous information derived from sleep and behavioral data.
To this end, we leverage \emph{sleep summary} features extracted from the previous night, together with smartwatch-derived \emph{step count} and \emph{activity statistics} from the previous day.
These are complemented by static \emph{demographic variables} (i.e., age, gender, BMI) and \emph{calendar-based} contextual features, such as the day of the week and holidays.


Moreover, to capture temporal dependencies, we construct a set of \emph{historical features} for each variable by aggregating past observations over multiple time scales.
Specifically, we consider: (i) \textit{lagged features}, representing daily values over the previous $N=3$ days; (ii) \textit{rolling statistics}, computed over fixed windows of $3$ and $7$ days to capture short- and medium-term trends; and (iii) \textit{exponentially weighted moving} (EWM) statistics, computed over the entire subject history to emphasize recent observations while retaining long-term context.
We adopt two EWM configurations with half-life parameters set to $3$ and $7$ days, analogous to the rolling windows, to model different temporal memory scales.


The \textbf{Sleep Duration} and \textbf{Sleep AHI} tasks follow a consistent feature construction pipeline, enabling a direct comparison across different levels of observability.
In these settings, we use multimodal smartwatch-derived signals from the previous day, including step count, activity levels, HR, BR, stress, and SpO$_2$, together with \emph{sleep-related} information from the previous night. Static and contextual predictors, as well as the same set of historical features, are also included.

Across all tasks, we restrict smartwatch data to daytime measurements (between $8$ A.M. and $8$ P.M.) to exclude samples potentially associated with sleep periods.
From these data, we compute a set of descriptive statistics (e.g., minimum, maximum, mean, median, and coverage), along with modality-specific features such as resting heart rate and the proportion of time spent in sedentary, active, and highly active states.
For sleep data, we focus on nocturnal events within the complementary time window and include all aggregated metrics provided by sleep summaries.
We further derive additional features capturing sleep structure and quality, including \emph{sleep stage ratios}, \emph{timing indicators} (e.g., midpoint hour), and fragmentation measures.

For rolling windows, we compute a comprehensive set of $24$ statistical descriptors.
However, not all of these are suitable for EWM representations. Since EWM assigns exponentially decaying weights to past observations, order-based or boundary-dependent statistics (e.g., minimum, maximum, and coverage) become less meaningful or ill-defined. Therefore, for EWM features we adopt a reduced yet informative set of statistics, including the mean, standard deviation, and the average absolute difference between consecutive observations. A summary of all feature types is reported in Table~\ref{tab:ML_features}.

\begin{table}[t]
\centering
\small
\caption{Dataset statistics before and after preprocessing.}
\begin{tabular}{l rrr rrr}
\toprule
\textbf{Task} & \multicolumn{3}{c}{\textbf{Before}} & \multicolumn{3}{c}{\textbf{After}} \\
\cmidrule(lr){2-4} \cmidrule(lr){5-7}
 & \textbf{Subjects} & \textbf{Observations} & \textbf{Features} & \textbf{Subjects} & \textbf{Observations} & \textbf{Features} \\
\midrule
Activity Levels & $64$ & $10241$ & $2731$ & $53$ & $5798$ & $101$ \\
Sleep Duration  & $64$ & $9292$  & $2253$ & $51$ & $4723$ & $102$ \\
Sleep AHI       & $63$ & $8015$  & $2253$ & $47$ & $4382$ & $139$ \\
\bottomrule
\end{tabular}
\label{tab:ML_dataset_stats}
\end{table}

\begin{table}[t]
\centering
\small
\caption{Hyperparameter search space and total number of configurations for each model.}
\begin{tabular}{llll}
\hline
\textbf{Model} & \textbf{Hyperparameter} & \textbf{Search Space} & \textbf{\# Configs} \\
\hline

\multirow{3}{*}{LR}
& $C$ & [0.01, 0.1, 1.0, 10.0] & \multirow{3}{*}{8} \\
& penalty & [l2] &  \\
& solver & [lbfgs, saga] &  \\
\hline

\multirow{5}{*}{RF}
& n\_estimators & [100, 300] & \multirow{5}{*}{48} \\
& max\_depth & [None, 10, 20] &  \\
& min\_samples\_split & [2, 5] &  \\
& min\_samples\_leaf & [1, 2] &  \\
& max\_features & [sqrt, log2] &  \\
\hline

\multirow{4}{*}{MLP}
& hidden\_layer\_sizes & [(64), (128), (64,32)] & \multirow{4}{*}{24} \\
& activation & [relu, tanh] &  \\
& alpha & [1e-4, 1e-3] &  \\
& learning\_rate\_init & [1e-3, 1e-2] &  \\
\hline

\end{tabular}
\label{tab:ML_hyperparams}
\end{table}

\subsubsection{AI pipeline}


Following feature extraction, we design a structured AI pipeline to enable a robust evaluation of cross-subject generalization across the 3 prediction tasks.
As a first step, we apply a conservative preprocessing strategy to ensure that models are trained exclusively on reliable observations.
Specifically, we remove features with a high proportion of missing values ($>30\%$), and subsequently discard samples containing missing entries.
While this choice reduces the overall dataset size, it avoids introducing additional assumptions related to data imputation, which is beyond the scope of this work.
This approach ensures that performance differences across tasks can be attributed to their intrinsic observability, rather than to preprocessing artifacts.


We then perform correlation-based feature selection to retain the most informative variables.
Features exhibiting weak absolute correlation with the target are discarded, resulting in a more compact and stable representation that reduces the risk of overfitting.
To further ensure statistical reliability, we retain only subjects with a sufficient number of daily observations ($>30$).
Finally, all features are standardized to mitigate inter-subject variability and improve comparability across individuals.
Dataset statistics, including the number of subjects, observations, and features before and after preprocessing, are reported in Table~\ref{tab:ML_dataset_stats}.


In our experiments, we evaluate three standard models, namely Logistic Regression (LR), Random Forest (RF), and Multilayer Perceptron (MLP), under a consistent experimental protocol.
Model performance is assessed using a nested cross-validation framework. In the outer loop, we adopt Leave-One-Subject-Out (LOSO) cross-validation to obtain an unbiased estimate of generalization performance on unseen individuals, iteratively holding out each subject as the test set. Within the inner loop, we perform subject-wise K-group cross-validation (with $K=4$) for hyperparameter tuning and model selection via grid search. The corresponding search spaces are reported in Table~\ref{tab:ML_features}.

We use the macro F1-score as the primary optimization metric to account for class imbalance. Class distributions for each task are reported in Figure~\ref{fig:activity_class_dist}. 
After selecting the optimal hyperparameters, each model is retrained on the full inner training set and evaluated on the held-out subject. This procedure ensures that the reported performance reflects true subject-level generalization. For each model and task, we report the distribution of macro F1-score, as well as standard Accuracy (Acc) and Balanced accuracy (BA) across the testing subjects.

\begin{figure}[t]
    \centering

    \begin{subfigure}[b]{0.32\textwidth}
        \centering
        \includegraphics[width=\textwidth]{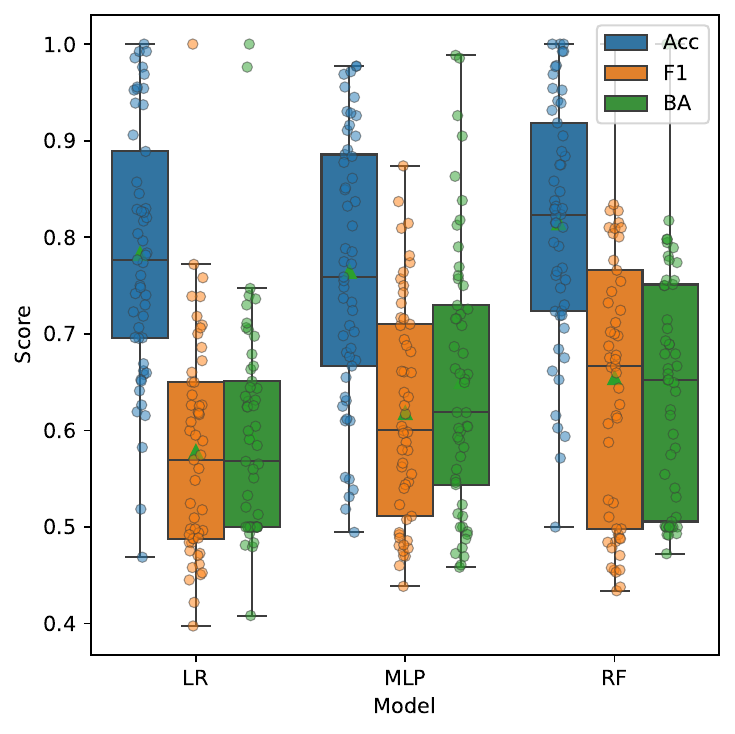}
        \caption{Activity Levels}
        \label{fig:activity_perf}
    \end{subfigure}
    \hfill
    \begin{subfigure}[b]{0.32\textwidth}
        \centering
        \includegraphics[width=\textwidth]{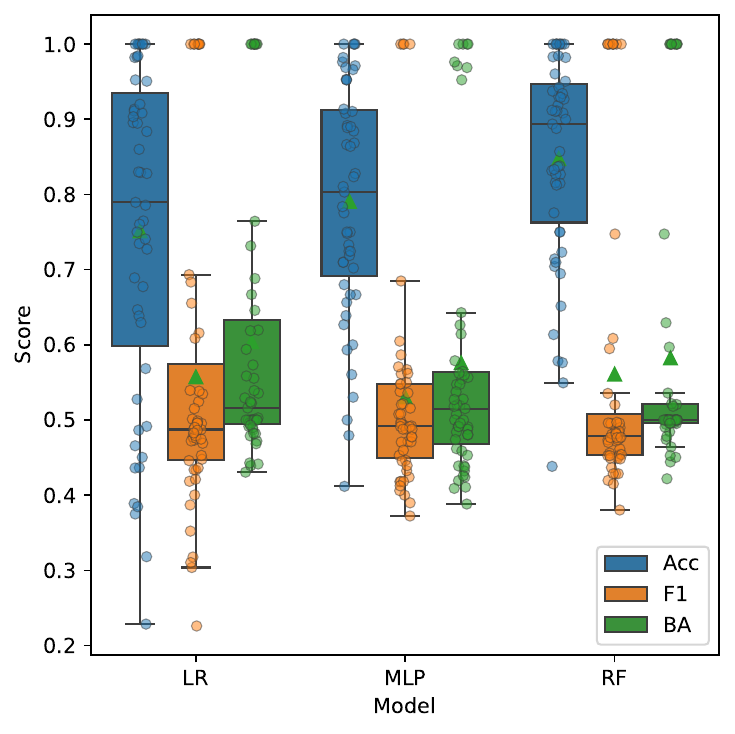}
        \caption{Sleep duration}
        \label{fig:sleep_dur_perf}
    \end{subfigure}
    \hfill
    \begin{subfigure}[b]{0.32\textwidth}
        \centering
        \includegraphics[width=\textwidth]{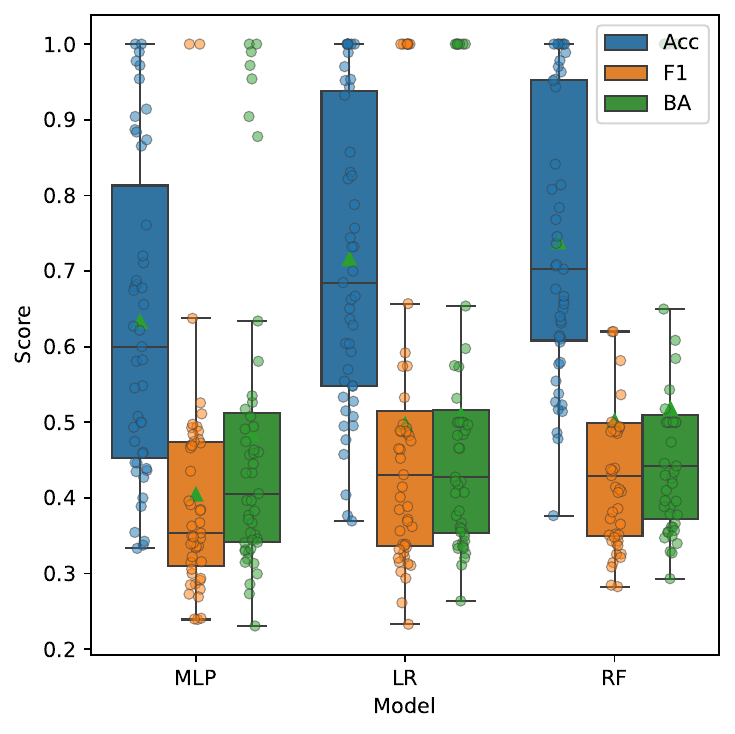}
        \caption{Sleep AHI}
        \label{fig:ahi_perf}
    \end{subfigure}

    \caption{Cross-subject performance distribution across predictive tasks.}
    \label{fig:ml_exp_performance}
\end{figure}

\subsubsection{Classification Results}


Figure~\ref{fig:ml_exp_performance} shows the classification performance of the considered models across the three target tasks.
For completeness, we also include accuracy, although this metric is less informative when considered in isolation, particularly in the presence of class imbalance.
To provide a meaningful reference, we compare all models against two simple baselines: a \emph{Random Guesser} (\emph{RG}), which assigns labels uniformly at random, and a \emph{Majority-class Classifier} (\emph{MC}), which always predicts the most frequent class.
Overall, the results reveal a clear gradient of predictability across tasks, consistent with their level of observability.

For the \textbf{Activity Levels} task, all models achieve strong and consistent improvements over the baselines, indicating that the target is clearly learnable and well-aligned with the available sensing modalities.
RF attains the best overall performance, reaching an average BA of $65.3\%$ (corresponding to a $+15.3\%$ improvement over baselines) and a macro-F1 of $65.4\%$, with gains of $+27.7\%$ over the majority classifier and $+15.4\%$ over RG.
MLP follows closely (BA $65.0\%$, macro-F1 $61.8\%$), while LR also shows competitive performance (BA $59.4\%$, macro-F1 $57.8\%$).
The consistency of these gains across different model families suggests that performance is not driven by a specific architecture, but rather by the strength of the underlying signal.
In terms of accuracy, RF achieves the highest value ($81.5\%$), significantly exceeding MC; importantly, these gains are aligned with improvements in class-balanced metrics, indicating genuine discriminative capability rather than bias toward the dominant class. While overall performance is solid, residual variability across subjects (standard deviation between $11\%$ and $15\%$) highlights that predictability is not uniform, leaving room for further improvements in generalization.

For the \textbf{Sleep Duration} task, all models again outperform both baselines in terms of class-balanced metrics, although performance is generally lower and more variable, reflecting the reduced observability of the target.
LR achieves the strongest balanced performance (BA $60.4\%$, macro-F1 $55.7\%$), followed closely by RF (BA $58.2\%$, macro-F1 $56.1\%$) and MLP (BA $57.6\%$, macro-F1 $53.0\%$).
These results indicate that meaningful predictive signals are present, even though the relationship between wearable data and sleep duration is less direct.
In contrast, accuracy is less informative in this setting due to pronounced class imbalance: while RF achieves the highest value ($84.8\%$), improvements over MC are limited, and LR and MLP remain close to it, confirming that accuracy is largely driven by class proportions.
A key characteristic of this task is the substantial cross-subject variability, with standard deviations ranging from approximately $13\%$ to $22\%$.
Notably, the distributions are skewed toward higher values, with several subjects exhibiting near-perfect BA and F1 scores.
This suggests that certain individuals are inherently easier to predict, likely due to stronger signal-to-label alignment, while others remain significantly more challenging, contributing to the observed variability. 

Finally, the \textbf{Sleep AHI} task represents the most challenging setting, as it involves the prediction of a clinically defined outcome characterized by weak coupling with wearable-derived signals.
Despite this, all models substantially outperform both baselines on the most informative metrics.
In particular, BA exceeds $48\%$ for all models, with RF achieving the best performance (BA $51.8\%$), corresponding to a $+18.5\%$ improvement over RG. Similarly, macro-F1 increases from $33.3\%$ (RG) and $24.4\%$ (MC) to approximately $50\%$ for LR and RF, confirming that the models capture meaningful patterns beyond trivial baselines.
Accuracy comparisons are again less straightforward due to class imbalance; nevertheless, all models surpass the majority classifier (Acc $57.7\%$), with RF reaching $73.8\%$ and LR $71.6\%$, while MLP achieves lower performance ($63.3\%$).
Interestingly, LR performs competitively with RF in terms of macro-F1, suggesting that even linear models can capture relevant structure in this task.
However, performance variability across subjects is particularly high (standard deviation $\approx 18$–$23\%$), indicating that the task remains inherently difficult.
As in the previous task, some subjects exhibit near-perfect performance, suggesting that prediction is highly dependent on individual-specific factors such as data quality, behavioral consistency, or class separability.

In conclusion, all the results demonstrate clear improvements over baselines and confirm the presence of exploitable signals. However, achieving robust and consistent performance across individuals remains challenging, which can be partly attributed to the high temporal stability of the underlying datas. The limited intra-subject variability reduces the availability of discriminative patterns, making it more difficult for models to generalize reliably across different individuals.

\subsubsection{Explainability}

\begin{figure}[t]
    \centering
    \includegraphics[width=.8\textwidth]{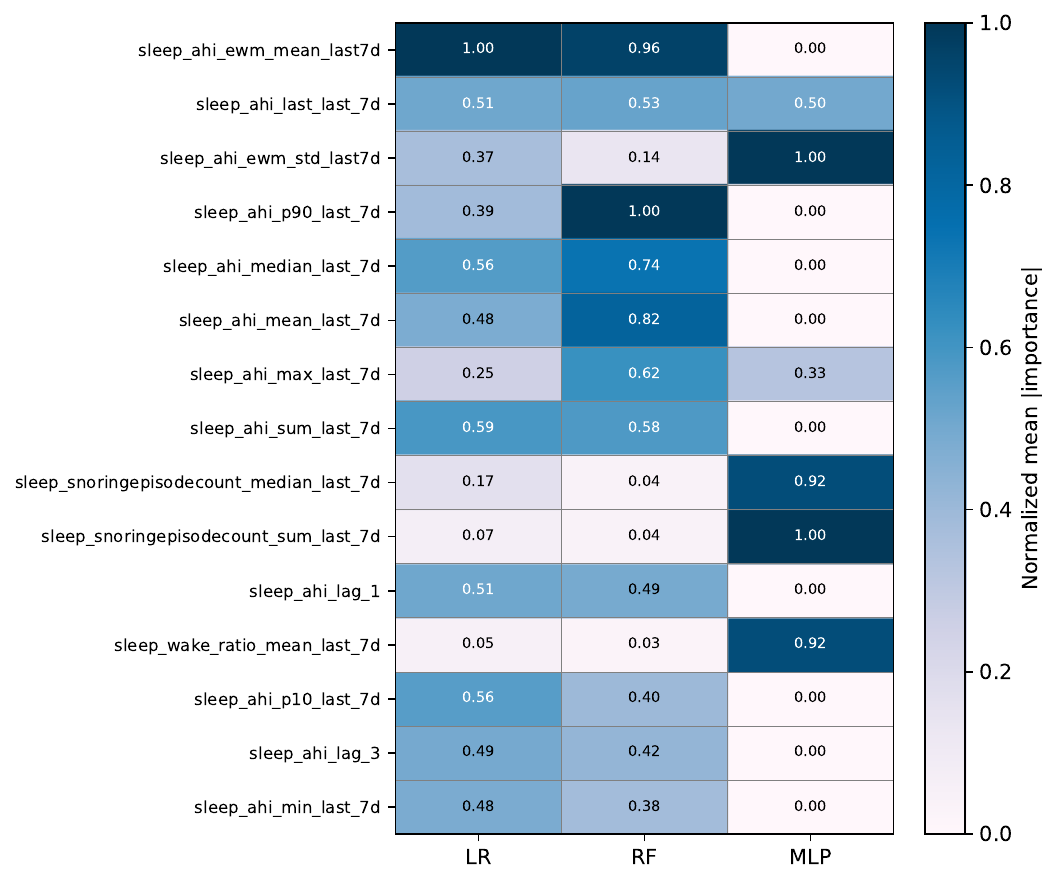}
    \caption{Comparative heatmap of the top $15$ most important features across LR, RF, and MLP models for the sleep AHI prediction task.}
    \label{fig:heatmap_features}
\end{figure}
To better understand the factors driving model predictions and assess the relevance of the input features, we performed an explainability analysis.
For each model, we computed feature attribution scores on the corresponding test set, using the most appropriate method for each model family.
For LR, we directly use the learned coefficients as a measure of feature importance, while for the remaining models, we rely on SHapley Additive exPlanations (SHAP)~\cite{SHAP} to estimate feature contributions.
In particular, we use TreeSHAP~\cite{DBLP:journals/corr/abs-1802-03888} to obtain exact Shapley value computations for RF, while KernelSHAP~\cite{SHAP} is employed to approximate feature attributions for the MLP.
For each model, we then computed the mean absolute importance across test subjects, allowing us to rank features and assess their consistency across model families.

Figure~\ref{fig:heatmap_features} presents the resulting feature importance heatmap for the \textbf{Sleep AHI} task, shown as a representative example among the three analyzed settings.
The corresponding visualizations for the \textbf{Activity Levels} and \textbf{Sleep Duration} tasks are shown in Figure~\ref{fig:additional_heatmaps_xai} in the Appendix for completeness.
To enable a direct comparison across models, feature importance scores are first normalized within each model, and then aggregated to compute a global ranking of the top-$15$ features. The heatmap displays these features ordered by decreasing global importance, highlighting both shared and model-specific patterns.
As we can note, the highest agreement is observed between LR and RF, which share $11$ out of the top-$15$ features ($73.3\%$).
In contrast, the MLP exhibits substantially lower agreement, with only $2$ features shared with either LR or RF ($13.3\%$). This discrepancy may partially stem from the approximate nature of KernelSHAP, which can introduce additional variability in feature attribution for non-linear models.

A similar analysis across the other tasks reveals a consistent pattern, although with varying degrees of agreement.
For the \textbf{Sleep Duration} task, LR and RF still show the strongest overlap, albeit reduced to $6$ shared features ($40.0\%$), while the MLP again exhibits limited agreement (between $6.7\%$ and $13.3\%$).
In contrast, for the \textbf{Activity Levels} task, the agreement is more evenly distributed across all model pairs, with $7$ shared features between RF and MLP ($46.7\%$), $5$ between LR and RF ($33.3\%$), and $4$ between LR and MLP ($26.7\%$).
This more balanced overlap suggests the absence of a dominant pairwise agreement, and instead indicates a broader consensus across models, which tend to focus on similar sets of informative features.

To provide a more systematic comparison, Table~\ref{tab:jaccard_index} reports the \emph{Jaccard Index} for each model pair across tasks, quantifying the overlap between feature sets as the ratio between their intersection and union.
The table confirms the trends observed in the heatmaps, highlighting substantial variability in agreement depending on the task.
In particular, the \textbf{Sleep AHI} task exhibits a strong alignment between LR and RF ($57.9\%$), while agreement with MLP remains minimal.
The \textbf{Sleep Duration} task shows moderate agreement ($25.0\%$) limited to LR and RF, whereas all other pairs exhibit very low similarity.
Finally, the \textbf{Activity Levels} task is characterized by a more distributed overlap, with no clearly dominant pair, reflecting a more consistent cross-model consensus.
Overall, these results suggest that agreement across models is higher in tasks where the underlying signal is more structured or strongly aligned with the feature space, while lower agreement may indicate either weaker observability or increased model sensitivity to different aspects of the data.

Another key aspect emerging from this analysis is that the top-ranked features are consistently \textbf{historical} in nature, particularly those derived from medium- to long-term temporal windows (e.g., one week).
This indicates that the learning process is primarily driven by temporally aggregated predictors, which capture behavioral regularities unfolding over extended periods.
Notably, this trend is consistent across all tasks, suggesting that the underlying phenomena exhibit strong temporal dependencies that cannot be fully captured by short-term observations alone.

These findings further emphasize the importance of longitudinal monitoring, as it enables both improved predictive performance and the extraction of stable and informative patterns that would otherwise remain hidden in short-term data.

\begin{table}[t]
\centering
\caption{Jaccard Index ($\%$) between model pairs across tasks.}
\label{tab:jaccard_index}
\begin{tabular}{lccc}
\hline
 & \multicolumn{3}{c}{\textbf{Model Pair}} \\
\cmidrule(lr){2-4}
\textbf{Task} & \textbf{LR--RF} & \textbf{RF--MLP} & \textbf{MLP--LR} \\
\hline
Activity Levels & 20.0 & 30.4  & 15.4 \\
Sleep Duration & 25.0 & 7.1 & 7.1  \\
Sleep AHI & 57.9 & 3.5 & 7.1 \\
\hline
\end{tabular}
\end{table}

\section{Conclusions and Future work}
\label{sec:conclusions}

In this work, we presented a longitudinal multimodal study of older adults conducted in real-world conditions, combining wearable sensing, mobile data collection, and periodic clinical assessments over an extended observation period.
The resulting dataset provides a rich representation of daily life collected into-the-wild, capturing both short-term behavioral variability and long-term health-related trajectories.
In this way, this study contributes to bridge the gap between controlled experimental settings and real-world health monitoring.

Beyond the dataset itself, we proposed a unified analysis framework based on the concept of observability, investigating how the alignment between wearable-derived signals and target variables affects predictive performance.
Through a set of machine learning experiments that span tasks with increasing levels of abstraction, ranging from activity prediction to clinically defined sleep outcomes, we showed that predictive performance is strongly influenced by the intrinsic observability of the target.
Although highly observable behavioral signals can be reliably inferred, more abstract or clinically grounded constructs remain challenging, despite the presence of meaningful predictive patterns.
These findings provide a principled lens for interpreting both the potential and limitations of wearable-based inference, particularly in longitudinal settings.

Importantly, this work specifically targets an older population, which remains underrepresented in existing wearable datasets and modeling efforts.
Our results highlight both the feasibility and the challenges of predictive modeling in this context, where inter-subject variability, heterogeneous behavioral patterns, and complex physiological dynamics require careful modeling choices and evaluation protocols.
In this sense, our study contributes not only a dataset, but also methodological insights that are directly relevant for the design of predictive systems tailored to aging populations.

Looking ahead, several research directions emerge from this work.
First, while our experiments rely on classical machine learning models with engineered features, future work could explore more advanced time-series modeling approaches, including \emph{deep learning architectures} specifically designed for sequential data, such as \emph{recurrent}, \emph{convolutional}, or \emph{transformer-based} models~\cite{10.1145/3558884.3558895}.
These methods could better capture temporal dependencies and long-range patterns that are only partially represented through lagged and rolling features.
At the same time, hybrid approaches combining feature-based representations with temporal models may provide a promising trade-off between interpretability and predictive power. Second, the longitudinal nature of the dataset opens the door to alternative modeling paradigms beyond per-sample prediction.
In particular, \emph{feature-level longitudinal modeling} and \emph{subject-centric representations} could be leveraged to better capture individual behavioral trajectories, enabling personalized modeling strategies and improving cross-subject generalization~\cite{Onnela2016}.
This is especially relevant in older populations, where variability across individuals is often more pronounced and clinically meaningful.
Finally, our dataset and findings contribute to the emerging area of \emph{Wearable Behavioural Foundation Models} (\emph{WBFMs})~\cite{10.1145/3749479, wu2026wearablefoundationmodelsstatic}, which aim to learn general-purpose representations from large-scale, multimodal behavioral data.
By providing a longitudinal, multimodal, and clinically anchored dataset collected into-the-wild, this work offers a valuable resource for pretraining and evaluating such models.
Future research could explore self-supervised and representation learning approaches to extract transferable embeddings, enabling more robust and scalable inference across tasks and populations.

In conclusion, this work highlights both the opportunities and the inherent challenges of longitudinal multimodal sensing for health monitoring in older adults.
By combining dataset design, empirical analysis, and predictive modeling, we provide a step toward more realistic, interpretable, and generalizable wearable-based health inference systems, while outlining a path for future research at the intersection of sensing, machine learning, and digital health.

\bibliographystyle{plain}
\bibliography{main}

@ARTICLE{Garg2018-zn,
  title     = "Hypertension in older adults: What is the target blood pressure?",
  author    = "Garg, Anu and Messinger-Rapport, Barbara J",
  journal   = "Cleve. Clin. J. Med.",
  publisher = "Cleveland Clinic Journal of Medicine",
  volume    =  85,
  number    =  3,
  pages     = "193--195",
  month     =  mar,
  year      =  2018,
  language  = "en",
  doi       = " https://doi.org/10.3949/ccjm.85a.17064"
}

@article{10.1371/journal.pmed.1001953,
    doi = {10.1371/journal.pmed.1001953},
    author = {Piwek, Lukasz AND Ellis, David A. AND Andrews, Sally AND Joinson, Adam},
    journal = {PLOS Medicine},
    publisher = {Public Library of Science},
    title = {The Rise of Consumer Health Wearables: Promises and Barriers},
    year = {2016},
    month = {02},
    volume = {13},
    url = {https://doi.org/10.1371/journal.pmed.1001953},
    pages = {1-9},
    number = {2},
}

@article{Winter2014,
  title={BMI and all-cause mortality in older adults: a meta-analysis},
  author={Winter, Jennifer E. and MacInnis, Robyn J. and Wattanapenpaiboon, Natalie and Nowson, Caryl A.},
  journal={Obesity},
  volume={22},
  number={1},
  pages={--},
  year={2014},
  doi={10.1002/oby.21612}
}

@article{APA,
  title={Promoting well-being in old age: The psychological benefits of two training programs of adapted physical activity},
  author={Delle Fave, Antonella and Bassi, Marta and Boccaletti, Elena S and Roncaglione, Carlotta and Bernardelli, Giuseppina and Mari, Daniela},
  journal={Frontiers in psychology},
  volume={9},
  pages={828},
  year={2018},
  publisher={Frontiers Media SA}
}

@article{6MWT,
author = {Paul L Enright},
title ={The Six-Minute Walk Test},
journal = {Respiratory Care},
volume = {48},
number = {8},
pages = {783-785},
year = {2003},
doi = {10.4187/respcare.03480783}
}

@article{MMSE,
  title={The mini-mental state examination: a comprehensive review},
  author={Tombaugh, Tom N and McIntyre, Nancy J},
  journal={Journal of the American Geriatrics Society},
  volume={40},
  number={9},
  pages={922--935},
  year={1992},
  publisher={Wiley Online Library}
}

@article{PSQI,
  title={The Pittsburgh sleep quality index as a screening tool for sleep dysfunction in clinical and non-clinical samples: A systematic review and meta-analysis},
  author={Mollayeva, Tatyana and Thurairajah, Pravheen and Burton, Kirsteen and Mollayeva, Shirin and Shapiro, Colin M and Colantonio, Angela},
  journal={Sleep medicine reviews},
  volume={25},
  pages={52--73},
  year={2016},
  publisher={Elsevier}
}

@article{DASS-21,
  title={The short-form version of the Depression Anxiety Stress Scales (DASS-21): Construct validity and normative data in a large non-clinical sample},
  author={Henry, Julie D and Crawford, John R},
  journal={British journal of clinical psychology},
  volume={44},
  number={2},
  pages={227--239},
  year={2005},
  publisher={Wiley Online Library}
}

@article{TUG,
  title={Timed Up and Go test and risk of falls in older adults: a systematic review},
  author={Beauchet, Olivier and Fantino, Bruno and Allali, Gilles and Muir, SW and Montero-Odasso, Manuel and Annweiler, C{\'e}dric},
  journal={The journal of nutrition, health \& aging},
  volume={15},
  number={10},
  pages={933--938},
  year={2011},
  publisher={Springer}
}

@article{IPAQ,
  title={Validity of the international physical activity questionnaire short form (IPAQ-SF): A systematic review},
  author={Lee, Paul H and Macfarlane, Duncan J and Lam, Tai Hing and Stewart, Sunita M},
  journal={International journal of behavioral nutrition and physical activity},
  volume={8},
  number={1},
  pages={115},
  year={2011},
  publisher={Springer}
}

@article{MNA,
  title={The Mini Nutritional Assessment (MNA) and its use in grading the nutritional state of elderly patients},
  author={Vellas, Bruno and Guigoz, Yves and Garry, Philip J and Nourhashemi, Fati and Bennahum, David and Lauque, Sylvie and Albarede, Jean-Louis},
  journal={Nutrition},
  volume={15},
  number={2},
  pages={116--122},
  year={1999},
  publisher={Elsevier}
}

@inproceedings{Mattingly2019Tesserae,
  title={The Tesserae Project: Large-Scale, Longitudinal, In Situ, Multimodal Sensing of Information Workers},
  author={Mattingly, S. and others},
  booktitle={Proceedings of the CHI Conference on Human Factors in Computing Systems},
  year={2019}
  
}

@article{Mundnich2020TILES,
  title={TILES-2018: A Longitudinal Physiological and Behavioral Dataset of Hospital Workers},
  author={Mundnich, K. and others},
  journal={arXiv preprint arXiv:2003.08474},
  year={2020}
}

@article{harari2016using,
  title={Using smartphones to collect behavioral data in psychological science: Opportunities, practical considerations, and challenges},
  author={Harari, Gabriella M and Lane, Nicholas D and Wang, Rui and Crosier, Benjamin S and Campbell, Andrew T and Gosling, Samuel D},
  journal={Perspectives on Psychological Science},
  volume={11},
  number={6},
  pages={838--854},
  year={2016},
  publisher={Sage Publications Sage CA: Los Angeles, CA}
}

@article{CORNET2018120,
title = {Systematic review of smartphone-based passive sensing for health and wellbeing},
journal = {Journal of Biomedical Informatics},
volume = {77},
pages = {120-132},
year = {2018},
issn = {1532-0464},
doi = {https://doi.org/10.1016/j.jbi.2017.12.008},
url = {https://www.sciencedirect.com/science/article/pii/S1532046417302782},
author = {Victor P. Cornet and Richard J. Holden}
}

@inproceedings{Wang2014StudentLife,
author = {Wang, Rui and Chen, Fanglin and Chen, Zhenyu and Li, Tianxing and Harari, Gabriella and Tignor, Stefanie and Zhou, Xia and Ben-Zeev, Dror and Campbell, Andrew T.},
title = {StudentLife: assessing mental health, academic performance and behavioral trends of college students using smartphones},
year = {2014},
isbn = {9781450329682},
publisher = {Association for Computing Machinery},
address = {New York, NY, USA},
url = {https://doi.org/10.1145/2632048.2632054},
doi = {10.1145/2632048.2632054},
booktitle = {Proceedings of the 2014 ACM International Joint Conference on Pervasive and Ubiquitous Computing},
pages = {3–14},
numpages = {12},
location = {Seattle, Washington},
series = {UbiComp '14}
}

@article{Xu2022GLOBEM,
author = {Xu, Xuhai and Liu, Xin and Zhang, Han and Wang, Weichen and Nepal, Subigya and Sefidgar, Yasaman and Seo, Woosuk and Kuehn, Kevin S. and Huckins, Jeremy F. and Morris, Margaret E. and Nurius, Paula S. and Riskin, Eve A. and Patel, Shwetak and Althoff, Tim and Campbell, Andrew and Dey, Anind K. and Mankoff, Jennifer},
title = {GLOBEM: Cross-Dataset Generalization of Longitudinal Human Behavior Modeling},
year = {2023},
issue_date = {December 2022},
publisher = {Association for Computing Machinery},
address = {New York, NY, USA},
volume = {6},
number = {4},
url = {https://doi.org/10.1145/3569485},
doi = {10.1145/3569485},
journal = {Proc. ACM Interact. Mob. Wearable Ubiquitous Technol.},
month = jan,
articleno = {190},
numpages = {34}
}

@article{10.1145/3328932,
author = {Saeed, Aaqib and Ozcelebi, Tanir and Lukkien, Johan},
title = {Multi-task Self-Supervised Learning for Human Activity Detection},
year = {2019},
issue_date = {June 2019},
publisher = {Association for Computing Machinery},
address = {New York, NY, USA},
volume = {3},
number = {2},
url = {https://doi.org/10.1145/3328932},
doi = {10.1145/3328932},
journal = {Proc. ACM Interact. Mob. Wearable Ubiquitous Technol.},
month = jun,
articleno = {61},
numpages = {30}
}

@article{Lifetrace2021,
author = {Bombassei De Bona, Francesco and C\^{a}mpanu, Ioana Andreea and Langheinrich, Marc and Gjoreski, Martin and Juravle, Georgiana},
title = {LIFETRACE: A Longitudinal Multimodal Dataset on Daily Physical Activity, Well-Being, and Habits},
year = {2025},
issue_date = {December 2025},
publisher = {Association for Computing Machinery},
address = {New York, NY, USA},
volume = {9},
number = {4},
url = {https://doi.org/10.1145/3770676},
doi = {10.1145/3770676},
journal = {Proc. ACM Interact. Mob. Wearable Ubiquitous Technol.},
month = dec,
articleno = {161},
numpages = {32}
}

@article{10.1145/3749502,
author = {Rossi, Alvise Dei and Marzorati, Davide and Gerosa, Tiziano and \v{S}vihrov\'{a}, Radoslava and Santini, Silvia and Faraci, Francesca},
title = {Unobtrusive Perceived Sleep Quality Monitoring in the Wild},
year = {2025},
issue_date = {September 2025},
publisher = {Association for Computing Machinery},
address = {New York, NY, USA},
volume = {9},
number = {3},
url = {https://doi.org/10.1145/3749502},
doi = {10.1145/3749502},
journal = {Proc. ACM Interact. Mob. Wearable Ubiquitous Technol.},
month = sep,
articleno = {77},
numpages = {26}
}

@inproceedings{Vaizman2017ExtraSensory,
  title={ExtraSensory: A Multimodal Dataset for Context Recognition in the Wild},
  author={Vaizman, Y. and Ellis, K. and Lanckriet, G.},
  booktitle={Proceedings of the ACM on Interactive, Mobile, Wearable and Ubiquitous Technologies},
  year={2017}
}

@ARTICLE{Liu2021-fo,
  title    = "Mobile health applications for older adults: a systematic review of interface and persuasive feature design",
  author   = "Liu, Na and Yin, Jiamin and Tan, Sharon Swee-Lin and Ngiam, Kee Yuan and Teo, Hock Hai",
  journal  = "J Am Med Inform Assoc",
  volume   =  28,
  number   =  11,
  pages    = "2483--2501",
  month    =  oct,
  year     =  2021,
  address  = "England",
  language = "en"
}

@misc{wu2026wearablefoundationmodelsstatic,
      title={Wearable Foundation Models Should Go Beyond Static Encoders}, 
      author={Yu Yvonne Wu and Yuwei Zhang and Hyungjun Yoon and Ting Dang and Dimitris Spathis and Tong Xia and Qiang Yang and Jing Han and Dong Ma and Sung-Ju Lee and Cecilia Mascolo},
      year={2026},
      eprint={2603.19564},
      archivePrefix={arXiv},
      primaryClass={cs.LG},
      url={https://arxiv.org/abs/2603.19564}, 
}

@article{10.1145/3749479,
author = {Qiu, Minghui and Weng, Cekai and Fan, Mingming and Wu, Kaishun},
title = {Towards Customizable Foundation Models for Human Activity Recognition with Wearable Devices},
year = {2025},
issue_date = {September 2025},
publisher = {Association for Computing Machinery},
address = {New York, NY, USA},
volume = {9},
number = {3},
url = {https://doi.org/10.1145/3749479},
doi = {10.1145/3749479},
journal = {Proc. ACM Interact. Mob. Wearable Ubiquitous Technol.},
month = sep,
articleno = {122},
numpages = {29}
}

@Article{Onnela2016,
author={Onnela, Jukka-Pekka
and Rauch, Scott L.},
title={Harnessing Smartphone-Based Digital Phenotyping to Enhance Behavioral and Mental Health},
journal={Neuropsychopharmacology},
year={2016},
month={Jun},
day={01},
volume={41},
number={7},
pages={1691-1696},
issn={1740-634X},
doi={10.1038/npp.2016.7},
url={https://doi.org/10.1038/npp.2016.7}
}

@inproceedings{10.1145/3558884.3558895,
author = {Augustinov, Gabriela and Nisar, Muhammad Adeel and Li, Fr\'{e}d\'{e}ric and Tabatabaei, Amir and Grzegorzek, Marcin and Sohrabi, Keywan and Fudickar, Sebastian},
title = {Transformer-Based Recognition of Activities of Daily Living from Wearable Sensor Data},
year = {2023},
isbn = {9781450396240},
publisher = {Association for Computing Machinery},
address = {New York, NY, USA},
doi = {10.1145/3558884.3558895},
booktitle = {Proceedings of the 7th International Workshop on Sensor-Based Activity Recognition and Artificial Intelligence},
articleno = {9},
numpages = {8},
location = {Rostock, Germany},
series = {iWOAR '22}
}

@article{sleep_duration,
  title={Sex differences in the association between sleep duration and frailty in older adults: evidence from the KNHANES study},
  author={Ha, Beomman and Han, Mijin and So, Wi-Young and Kim, Seonho},
  journal={BMC geriatrics},
  volume={24},
  number={1},
  pages={434},
  year={2024},
  publisher={Springer}
}

@article{SHAP,
  title={A unified approach to interpreting model predictions},
  author={Lundberg, Scott M and Lee, Su-In},
  journal={Advances in neural information processing systems},
  volume={30},
  year={2017}
}

@misc{WHO2020,
  title = {WHO Guidelines on Physical Activity and Sedentary Behaviour},
  author = {{World Health Organization}},
  year = {2020},
  publisher = {World Health Organization},
  url = {https://www.who.int/publications/i/item/9789240015128}
}

@misc{IFAPA,
  title = {What is Adapted Physical Activity?},
  author = {{International Federation of Adapted Physical Activity}},
  year = {2020},
  url = {https://ifapa.net/}
}

@article{Zhang2025ABPM,
  title   = {Ambulatory blood pressure monitoring, European guideline targets, and cardiovascular outcomes: an individual patient data meta-analysis},
  author  = {Zhang, D. Y. and An, D. W. and Yu, Y. L. and Melgarejo, J. D. and Boggia, J. and Martens, D. S. and Hansen, T. W. and Asayama, K. and Ohkubo, T. and Stolarz-Skrzypek, K. and Malyutina, S. and Casiglia, E. and Lind, L. and Maestre, G. E. and Wang, J. G. and Imai, Y. and Kawecka-Jaszcz, K. and Sandoya, E. and Rajzer, M. and Nawrot, T. S. and O'Brien, E. and Yang, W. Y. and Filipovsk{\'y}, J. and Graciani, A. and Banegas, J. R. and Li, Y. and Staessen, J. A. and {International Database of Ambulatory Blood Pressure in Relation to Cardiovascular Outcomes Investigators}},
  journal = {European Heart Journal},
  year    = {2025},
  volume  = {46},
  number  = {30},
  pages   = {2974--2987},
  doi     = {10.1093/eurheartj/ehafXXX}
}

@article{DBLP:journals/corr/abs-1802-03888,
  author       = {Scott M. Lundberg and
                  Gabriel G. Erion and
                  Su{-}In Lee},
  title        = {Consistent Individualized Feature Attribution for Tree Ensembles},
  journal      = {CoRR},
  volume       = {abs/1802.03888},
  year         = {2018},
  url          = {http://arxiv.org/abs/1802.03888},
  eprinttype   = {arXiv},
  eprint       = {1802.03888},
  bibsource    = {dblp computer science bibliography, https://dblp.org}
}

\appendix
\section*{Appendix}
\section{Additional Feature Importance Visualizations}
\begin{figure}[t]
    \centering

    \begin{subfigure}[t]{0.8\textwidth}
        \centering
        \includegraphics[width=\linewidth]{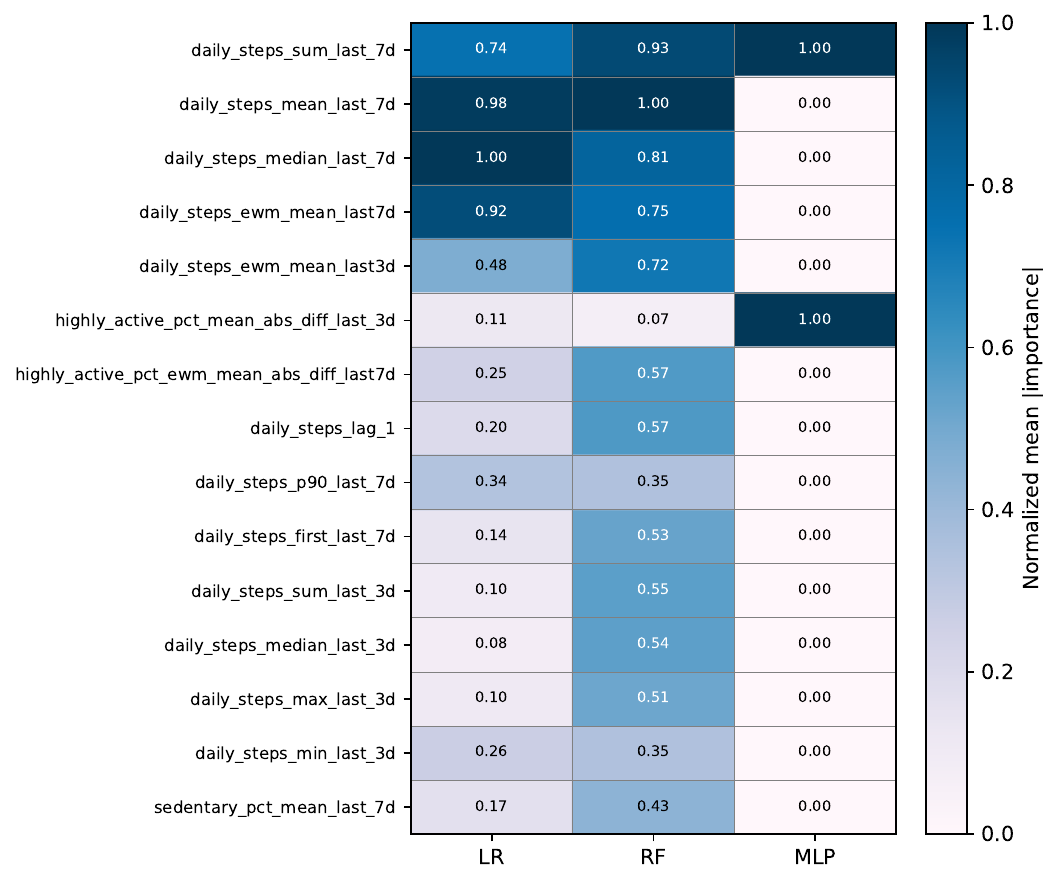}
        \caption{Activity Level}
        \label{fig:activity_level}
    \end{subfigure}

    \begin{subfigure}[t]{0.8\textwidth}
        \centering
        \includegraphics[width=\linewidth]{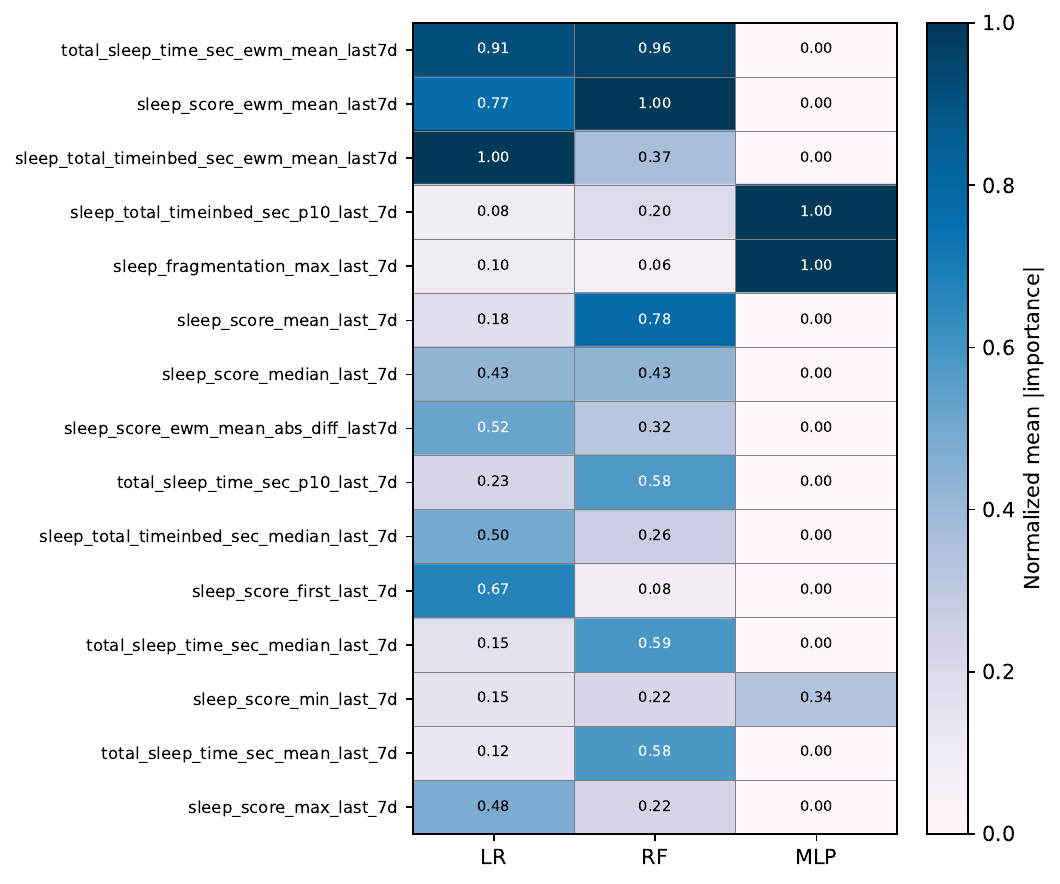}
        \caption{Sleep Duration}
        \label{fig:xai_sleep_duration}
    \end{subfigure}
    
    \caption{Comparative heatmap of the top $15$ most important features across LR, RF, and MLP models for the a) Activity Levels and b) Sleep Duration tasks.}
    \label{fig:additional_heatmaps_xai}
\end{figure}

For completeness, we report in this appendix the feature importance heatmaps for the \textbf{Activity Levels} (Figure~\ref{fig:activity_level}) and \textbf{Sleep Duration} (Figure~\ref{fig:xai_sleep_duration}) tasks, complementing the \textbf{Sleep AHI} analysis presented in the main text. These visualizations are obtained using the same methodology, where feature importance is estimated for each model, normalized within-model, and aggregated to derive a global ranking of the top-$15$ features.

Overall, the heatmaps confirm the trends discussed in the main paper. While the degree of agreement across models varies depending on the task, the most relevant predictors consistently correspond to \emph{historical features}, i.e., features that capture information from past observations over extended time horizons.
This pattern is particularly evident in the \textbf{Activity Levels} task, where a relatively higher agreement across models is associated with a shared focus on temporally aggregated information. Similarly, in the \textbf{Sleep Duration} task, although the overlap across models is more limited, the top-ranked features still predominantly reflect historical information rather than instantaneous measurements.

In conclusion, these results reinforce the key finding that longitudinal information plays a central role in model performance across all tasks. The consistent prominence of historical features suggests that the relevant behavioral and physiological patterns evolve over time and cannot be fully captured by short-term observations alone.

\end{document}